\documentclass[10pt,twocolumn,letterpaper]{article}

\usepackage{3dv}
\usepackage{times}
\usepackage{epsfig}
\usepackage{graphicx}
\usepackage{amsmath}
\usepackage{amssymb}
\usepackage{adjustbox}
\usepackage[dvipsnames]{xcolor}
\usepackage{float}
\usepackage{footnote}
\makesavenoteenv{tabular}
\makesavenoteenv{table}
\usepackage{booktabs}
\usepackage[skip=0.7em,labelfont=bf]{caption}
\usepackage{subcaption}
\usepackage{multirow}

\usepackage[pagebackref=true,breaklinks=true,letterpaper=true,colorlinks,bookmarks=false]{hyperref}

\threedvfinalcopy 

\def\threedvPaperID{128} 

\newcommand{\cA}{\mathcal{A}}
\ifthreedvfinal\fi

\def\formatparagraphtitle#1{#1.}
\makeatletter
\renewcommand\paragraph{\@startsection{paragraph}{4}{\z@}%
             {2ex plus 0.5ex minus 0.25ex}%
             {-1.0em}%
             {\reset@font\normalsize\bfseries\formatparagraphtitle}}
\makeatother

\begin{document}

\title{Learning to Infer Semantic Parameters for 3D Shape Editing}


\author{Fangyin Wei$^{1}$ \quad
Elena Sizikova$^{2}$ \quad
Avneesh Sud$^{3}$ \quad
Szymon Rusinkiewicz$^{1}$ \quad
Thomas Funkhouser$^{1,3}$\\
${}^{1}$ Princeton University  \quad\quad  ${}^{2}$ New York University  \quad\quad   ${}^{3}$ Google Research}

\maketitle

\begin{abstract}
Many applications in 3D shape design and augmentation require the ability to make specific edits to an object's semantic parameters (\eg, the pose of a person's arm or the length of an airplane's wing) while preserving as much existing details as possible.  We propose to learn a deep network that infers the semantic parameters of an input shape and then allows the user to manipulate those parameters. The network is trained jointly on shapes from an auxiliary synthetic template and unlabeled realistic models, ensuring robustness to shape variability while relieving the need to label realistic exemplars. At testing time, edits within the parameter space drive deformations to be applied to the original shape, which provides semantically-meaningful manipulation while preserving the details. This is in contrast to prior methods that either use autoencoders with a limited latent-space dimensionality, failing to preserve arbitrary detail, or drive deformations with purely-geometric controls, such as cages, losing the ability to update local part regions. Experiments with datasets of chairs, airplanes, and human bodies demonstrate that our method produces more natural edits than prior work.\footnote{https://github.com/weify627/learn-sem-param}
\end{abstract}

%
\vspace{-15pt}
\section{Introduction}

\begin{figure}[t!]
   \includegraphics[width=1\linewidth,trim={1cm 0.3cm 0.5cm 1cm}]{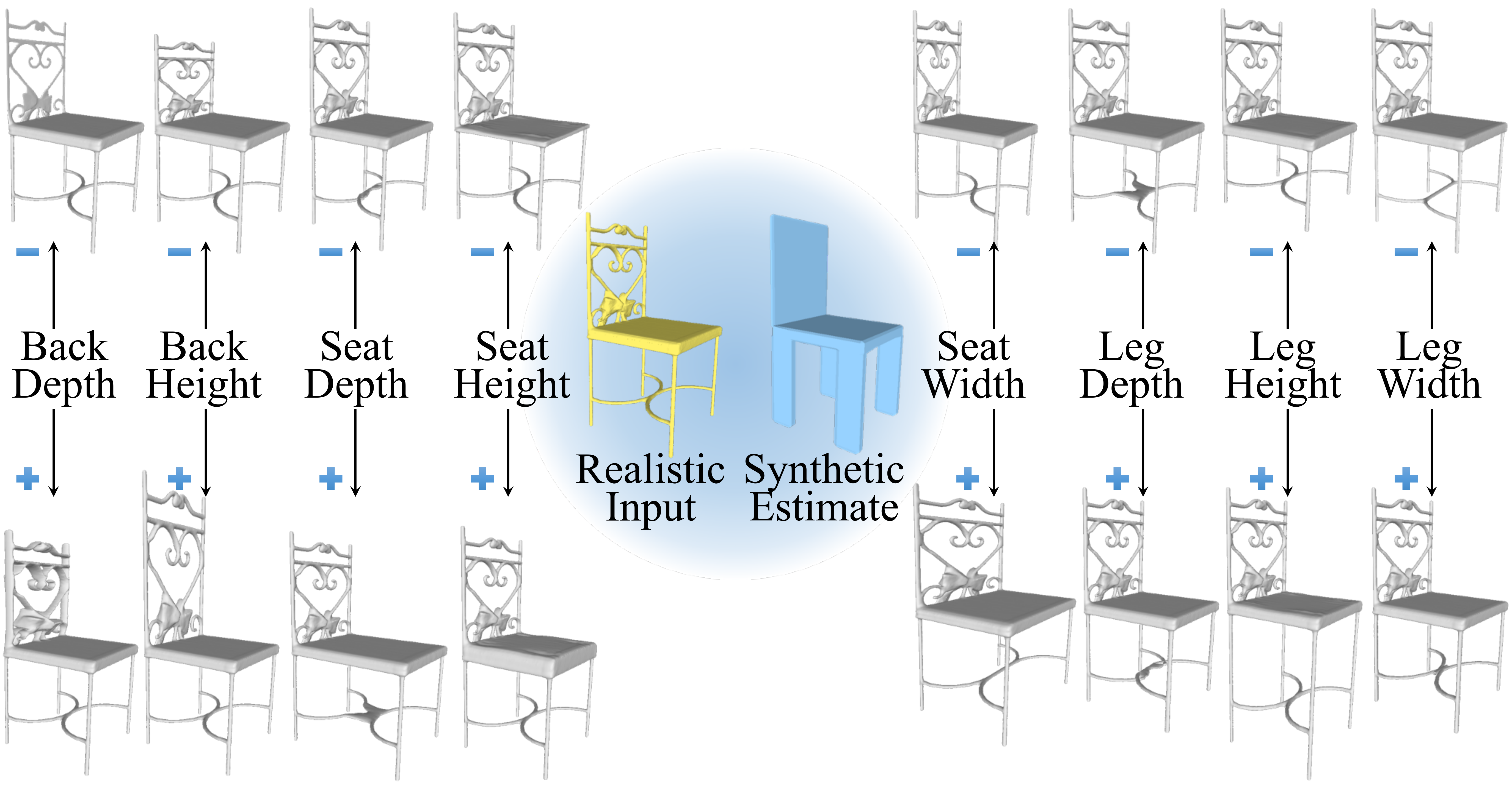}
   \vspace{-17pt}
   \caption{\textbf{Semantic editing}. Taking a realistic shape (yellow) as input, the proposed method allows for editing of different parts (gray). This is achieved by learning to infer the semantic parameters of an auxiliary synthetic shape (blue), without any labels from realistic shapes.}
   \vspace{-8pt}
\label{fig:teaser}
\end{figure}

The ability to perform semantically-meaningful manipulation of 3D shapes is crucial to many applications ranging from industrial design to dataset augmentation for 3D learning.  Although the space of possible manipulations is large, we focus on editing \emph{semantic parameters}, such as the angle of a person's leg or the height of a chair's seat (see Fig.~\ref{fig:teaser}).  For maximum control, we wish to allow adjusting these independently, such that changing one parameter preserves the others.  Moreover, we would like to preserve topology and fine detail throughout the input shape, both within and far away from the edited region.  Finally, to make our method broadly applicable, we would like to avoid dependency on large labeled datasets of 3D models or edits.

Traditional shape manipulation methods~\cite{ganapathi2018parsing,zheng2011component} first fit predefined handles (\eg, cages, primitives) to input shapes through optimization. The user then manipulates the template, which guides the deformation of the original shape. This approach has several disadvantages.  First, cages make it difficult or impossible to deform local regions, as opposed to the shape as a whole.  In addition, the algorithms that fit handles to an input shape can be initialization-sensitive.  Finally, this approach inherently focuses on \emph{geometric} parameters (\eg, cage control points), and establishing the link between them and semantic parameters (\eg, ``seat height'') would require significant training on large, labeled datasets.

Recently, learning-based methods for shape deformation have become the subject of active research. Some approaches~\cite{gao2019sdm,hao2020dualsdf,tan2018variational} build upon the autoencoder idea: an encoder network maps an input shape to a point in a latent ``shape space'', while a decoder network re-synthesizes a shape given a latent vector.  Using this idea for semantic editing requires training a disentangled encoder, which maps semantic parameters to certain latent dimensions.  This requires large, labeled training datasets.  More crucially, the fine-scale detail produced by the decoder is fundamentally limited by the latent space dimensionality, and by what detail was present in the training dataset.  This means that no resynthesis-based method will be able to preserve the detail present in arbitrary input shapes.

Another popular approach for 3D shape editing is to predict shape deformation~\cite{mehr2019disconet,wang20193dn,yifan2019neural}. When labeled edits (\eg, ``make the seat 50\% taller'') are available, a network can be trained with full supervision for a deformation task~\cite{yumer2015semantic,yumer2016learning}. However, collecting many such labeled edits is not trivial, and most learnable methods that predict deformation~\cite{wang20193dn,yifan2019neural} instead solve the task of source-to-target deformation. These methods can only globally deform a shape given another exemplar and do not support local edits with semantic awareness.   

In this work, we build upon several key ideas to edit 3D shapes in a semantically meaningful way. First, to allow for full detail preservation, we design our system around deformation and not re-synthesis.  The deformation model is flexible enough to allow for both global and local edits, enabling independent control over different semantic parameters.  Next, to infer the parameters of an input shape, we rely on an encoder network trained to embed each input into a ``semantic parameter space''.  Crucially, this is a many-to-one network that need not embed all details present in the input as a traditional ``shape space'' encoder.  We train this network using a combination of labeled synthetic shapes and \emph{unlabeled} realistic shapes.  We therefore gain the advantage of a semantically-parameterized latent space (whose dimensions are defined by template parameters) without the need for labeled realistic datasets.  Finally, at edit time, we use the learned encoder to extract semantic parameters from the input, and then deform the input based on how those parameters change the shape of the synthetic template.

There are several advantages to our method.  First, it learns only a semantically interpretable space that is relevant to the editing operations. This is in contrast to learning to encode the entire shape, which suffers from detail loss due to the fixed dimensionality of the latent space. 
Second, by abstracting realistic shapes into a shared semantic space with synthetic shapes, we sidestep the need for labels for realistic shapes.
Finally, the low-dimensional semantic space acts as a regularizer, making the encoder easier to learn. Experiments show that the model generalizes, enabling meaningful edits on shapes that fall out of training distributions. 

We test our proposed method on three classes covering both rigid and non-rigid shapes, for different editing tasks. For chairs and airplanes, we consider the application of anisotropic part deformation, driven by semantic labels that need not correspond to individual parts.  For example, we can have a ``leg width'' semantic parameter that controls all four legs simultaneously.  For human bodies, we show pose editing trained on a simple body model, which nevertheless generalizes to more realistic bodies.  Experiments show that the proposed method produces results that are consistent with the user's desired semantic edits while being more useful and more robust than prior work.

\begin{figure*}[t!]
   \includegraphics[width=1\linewidth,trim={0cm 0cm 0cm 0cm},clip]{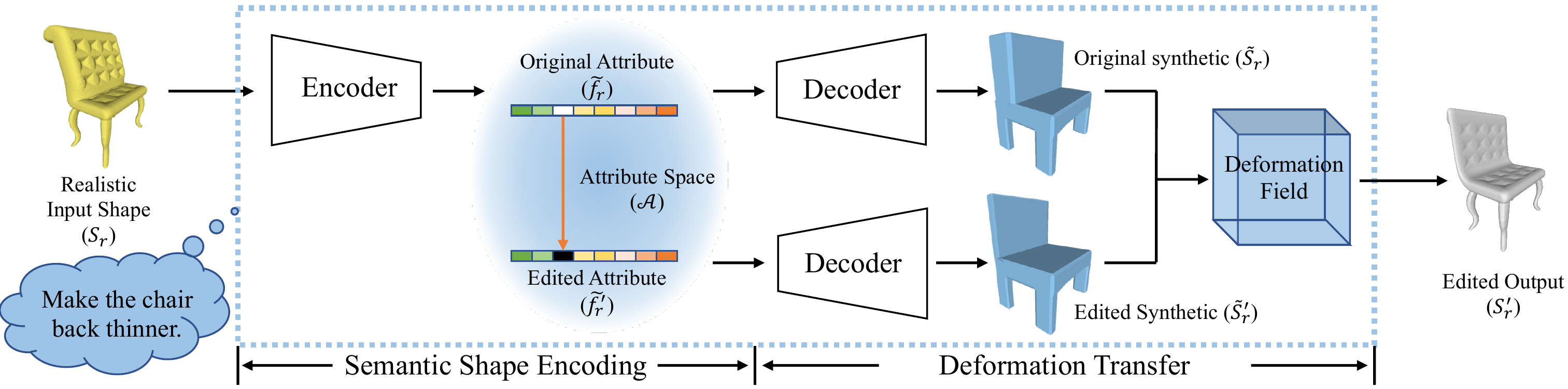}
  \vspace{-5pt}
   \caption{\textbf{Framework overview}. To achieve shape editing, we propose a framework that (a) learns a semantic parameter space for both realistic shapes $S_r$ and auxiliary synthetic shapes $S_s$ (Sec.~\ref{semantic_encoding}) and (b) transfers the deformation of synthetic shapes to realistic shapes (Sec.~\ref{deform}).}
\label{fig:overview}
\end{figure*}

\section{Related Work}

There is a long history of prior work on semantic shape editing \cite{mitra2014structure,xu2016data}. Relevant to our work, we discuss prior work in shape deformation.

\paragraph{Traditional Template-Based Shape Deformation}
Representative early efforts fit templates to input shapes through optimization. For example, Zheng et al.~\cite{zheng2011component} fit a “controller” (\eg, a cylinder) to every component decomposed from the input shape, then propagate user edits among components to guide input shape deformation. Their shape decomposition works for various man-made shapes, but is mostly based on geometry and is not semantically consistent across shapes. To achieve semantic consistency, Ganapathi-Subramanian et al.~\cite{ganapathi2018parsing} fit each input shape with a class-specific refined template.  However, such fitting algorithms are fragile, utilizing hand-crafted heuristics and thresholds. We provide a robust deep learning approach to semantic template fitting.


 \vspace{-2pt}
\paragraph{Supervised Style Attribute Learning} While traditional methods often involve manual supervision of templates and editing pipelines, learning-based approaches achieve more automatic editing with priors learned from a large amount of data. 
Yumer et al.~\cite{yumer2015semantic} learn a mapping from a shape to an attribute score to guide shape deformation, while Yumer et al.~\cite{yumer2016learning} use a deep network to directly learn a volumetric deformation field. However, such supervised methods are extremely data-hungry, whereas large-scale 3D shape datasets with annotated deformations are not currently available. 

\vspace{-2pt}
\paragraph{Deformation Retargeting with Global Structure} 
Another line of work~\cite{sumner2004deformation,Sung:2020,wang20193dn,yifan2019neural} sidesteps the need for dense annotations by instead focusing on deforming a source shape to globally resemble a target. Wang et al.~\cite{wang20193dn} learn to predict a coordinate offset for each vertex of the input using global shape descriptors, and Wang et al.~\cite{yifan2019neural} predict a 40-vertex cage to define the deformation field. While relying less on densely labeled data, these methods are not structure-aware and only allow for coarse shape deformation without precise control over local components. In concurrent work, Sung et al.~\cite{Sung:2020} learn class-specific deformations by projecting user-edited shapes onto a plausible learned shape space. Unlike our work, this method requires realistic source-target pairs for training, and does not guarantee shape preservation for unedited components.

\vspace{-2pt}
\paragraph{Deformation Retargeting with Learned Templates} To achieve more local editing, more recent work learns to locally deform a template and retarget the deformation to the input shape. Mehr et al.~\cite{mehr2019disconet} propose to learn a disconnected manifold from multiple auto-encoders with each deforming a different learned template. After training, they optimize over template deformation to fit a user edit, and the optimized template deformation is retargeted to edit the input shape. Although this approach creates a user-friendly interface, it does not support semantic edits and is sensitive to the robustness of latent space optimization.
Our method also deforms a realistic shape guided by a deformation field learned using synthetic shapes, and
  we can more specifically control different semantic components of the realistic shape, with more continuous and precise manipulation. Furthermore, while~\cite{mehr2019disconet} focuses on learning a disconnected manifold of shapes, we specifically advocate for learning only a small set of parameters for shape editing rather than learning to encode the entire shape.
 Meanwhile, there exist deformation retargeting approaches using domain-specific templates\,---\,statistical models as well as learned embeddings\,---\,for humans~\cite{Alldieck_2019,huang2020arch}. In contrast, our approach generalizes across rigid and non-rigid shapes.

\paragraph{Learning Representations for Shape Resynthesis}
%
A highly related field is learned shape reconstruction, which aims to reconstruct a 3D shape from an inferred latent code. Several recent shape reconstruction methods~\cite{gao2019sdm,hao2020dualsdf,liu2017interactive,mescheder2019occupancy,park2019deepsdf} allow for simple shape editing by manipulating the latent representation. Due to the limited scope of the latent space, these reconstruction algorithms can only reproduce shapes similar to ones observed in training, thus losing details after editing. Finally, the shape representation cannot be easily disentangled into semantic factors in the absence of labeled data~\cite{gao2019sdm,locatello2019challenging}. Therefore, we argue that for the purpose of shape editing, we should not encode the entire shape. Instead, it is sufficient to only encode the modes of shape variation corresponding to the desired edits, which can be efficiently represented with a small set of parameters. We analytically decode from the space of semantic parameters, which guides the deformation to be applied to the input.


\section{Approach}\label{sec:approach}

Our approach decomposes shape editing into two stages: semantic shape encoding and deformation transfer (Fig.~\ref{fig:overview}).
The original realistic shape is first passed into an encoder that infers its semantic parameters. Manipulation is then performed in the parameter space based on the target editing. Afterward, both original and edited semantic parameters are taken as input by an analytical decoder that outputs synthetic shapes reflecting the editing. The final deformation of the original input is guided by a deformation field defined by the predicted synthetic shapes. 

The main learning challenge in this approach is to train an encoder that can extract semantic parameters from arbitrary input shapes.  To address this challenge, we must answer two main questions: 1) how to allow a user to define semantic parameters, and 2) how to obtain training data for a network to learn to infer them.  

Our approach for the first question is to ask the model designer to create a synthetic template with semantic parameters defining a space of possible shapes (\eg, a template for chairs may have parameters controlling the seat depth, back height, leg length, \etc for a set of boxy primitives).  The templates are simple and produced once per object category, so the burden of creating them is small (a few minutes) and certainly far less than labeling a large dataset of examples.

Our approach for the second question is to procedurally build a semi-supervised training set with a mix of unlabeled realistic shapes and labeled synthetic shapes derived from the template.   We train the semantic encoder to predict the semantic parameters of the template examples and to match the shapes of all examples. Through joint training on realistic and synthetic examples, the network learns to infer semantic parameters for realistic shapes with only precise semantic supervision from synthetic data.

\begin{figure}[t!]
   \includegraphics[width=1\linewidth,trim={0cm 0cm 0cm 0cm},clip]{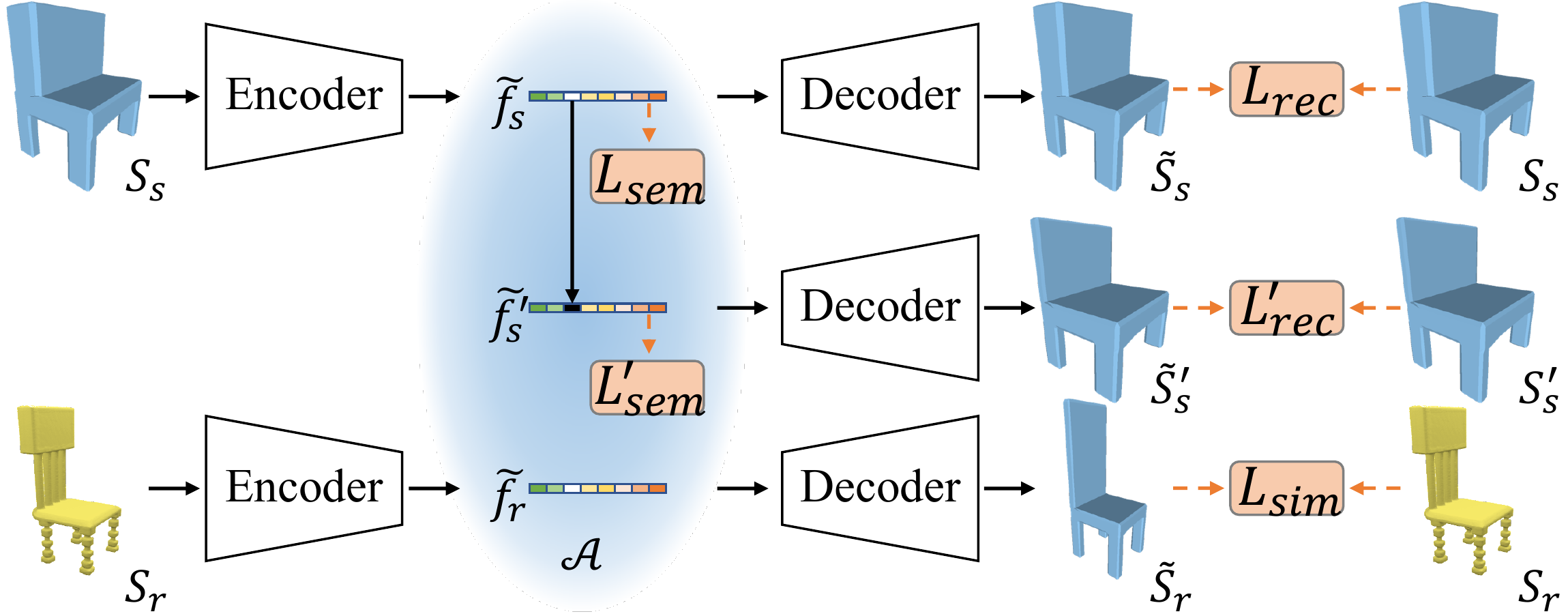}
   \caption{\textbf{Learning framework}. We learn a semantic space $\cA$ (blue) by jointly training on synthetic and realistic shapes. Synthetic shapes are supervised in both semantic and shape space, with one reconstruction branch (top) and one editing prediction (middle). Realistic shapes (bottom) are supervised in shape space only.}
\label{fig:learn}
\end{figure}

\vspace{-2pt}
\section{Method}
Our editing pipeline takes in a realistic input shape $S_r$ and generates its deformed version $S_r'$ defined by changes to semantic parameters (Fig.~\ref{fig:overview}).   To achieve this, we first use a trained encoder $E$ to infer semantic parameters $\tilde{f_r}$ of the input shape $S_r$.  Those parameters can be edited or optimized (depending on the application) to obtain $\tilde{f}_r'$.  We then use an analytical decoder to produce synthetic shapes $\tilde{S_r}$ and $\tilde{S}_r'$ from $\tilde{f_r}$ and $\tilde{f}_r'$ respectively.  Finally, the pair of synthetic shapes are used to define a deformation field $\mathcal{D}$ that is applied to the input shape $S_r$ to output shape $S_r'$.

\subsection{Semantic Shape Encoding}\label{semantic_encoding}

\paragraph{Learning Framework}

During training, we optimize the encoder to learn a semantic parameter space shared by both synthetic and realistic shapes. Specifically, in each mini-batch, we set half of the examples to be synthetic and half realistic. The encoded semantic parameters of synthetic shapes are passed to two branches. The first branch (top row in Fig.~\ref{fig:learn}) reconstructs the input shape from the encoded semantic feature $\tilde{f_s}(d=|\tilde{f_s}|)$. The second branch (middle row in Fig.~\ref{fig:learn}) predicts the edited shape from a new semantic feature $\tilde{f}_s'$ which is obtained by modifying the original feature $\tilde{f}_s$ according to a target edit. Both features are then decoded into synthetic shapes by the decoder. We show in Sec.~\ref{sec:ablation} that the second editing branch is essential for maintaining training stability and improving performance. Both branches are supervised by a semantic parameter loss $L_{sem}$ and a shape reconstruction loss $L_{rec}$. For each realistic shape (bottom row in Fig.~\ref{fig:learn}), the encoder estimates a semantic feature $\tilde{f_r}$, which is then decoded to a synthetic shape. A shape similarity loss $L_{sim}$ is used to make sure the synthetic shape is close to the realistic shape semantically, since no ground truth is available for realistic shapes.

\paragraph{Semantic Parameter Loss}
\def\dist{\mathit{dist}}
With the ground truth semantic parameters for synthetic shapes, we define
\[L_{sem}=\dist(f_s,\tilde{f_s}) \text{ and } L_{sem}'=\dist(f_s',\tilde{f}_s')\]
as semantic parameter losses for reconstruction and editing branches, respectively, where $\dist$ is a distance metric (\eg, L2 distance for scaling parameters, geodesic distance for rotational parameters). By training with supervision from synthetic shapes expressing a dense range of semantic parameter values, we learn to encode the full distribution of the space, which enables highly generalizable semantic edits by moving along dimensions in the space.
\paragraph{Shape Reconstruction Loss}
In the 3D shape space, since we know the exact vertex correspondence between input and predicted shapes, we have:
\[L_{rec}=\sum_{i=1}^n \bigl\|\tilde{v}_s^{(i)}-v_s^{(i)}\bigr\|^2 \text{ and } L_{rec}'=\sum_{i=1}^n \bigl\|\tilde{v}_s^{{\prime}(i)}-v_s^{{\prime}(i)}\bigr\|^2\] 
for the reconstruction and editing branches, respectively, where $n$ is the total number of vertices, and $v_s^{(i)}$ and $\tilde{v}_s^{(i)}$ are the vertices from the ground truth mesh and predicted synthetic mesh, respectively.

\paragraph{Shape Similarity Loss}
For realistic shapes, since no ground truth semantic parameters are available, we only impose a shape similarity loss in the 3D space between the output synthetic predictions and the input realistic shapes. We choose the shape similarity loss as the chamfer distance between randomly sampled point clouds from both shapes:
\[L_{sim}=\sum_{v\in \tilde{S}_r} \min_{u\in S_r}\|v-u\|^2+\sum_{u\in S_r} \min_{v\in \tilde{S}_r}\|u-v\|^2.\]

\paragraph{Overall Learning Objective}
The total loss $L$ is a weighted combination of all loss terms as described above:
\[
L = \alpha\,(L_{sem}+L_{sem}') + \beta\,(L_{rec}+L_{rec}') + \gamma \, L_{sim},
\]
where $\alpha,\beta,\gamma$ are the weights for individual terms.

\paragraph{Training Details} \label{sec:training_details}
The encoder can be flexible according to the input representation. For a point cloud input, we use the PointNet~\cite{qi2017pointnet}, which is composed of seven fully-connected layers interlaced with Batch Normalization and ReLU layers. The sizes of layers from bottom to top are (64, 128, 128, 256, $d$, $d$, $d$). For a mesh input, we use a graph CNN, which is composed of a fully-connected layer with output feature size 16, four FeaStConv~\cite{verma2018feastnet} layers with output feature size (32, 64, 96, 128), and one final fully-connected layer with a task-specific feature size $d$. All layers are interlaced with a Leaky ReLU layer. For non-rigid experiments, when all realistic input has the same mesh connectivity (\eg, DFAUST), we can use a mesh encoder. When the realistic shapes do not follow the same mesh structure (\eg, scanned data), we use a point encoder. The differentiable decoder takes as input the semantic parameters and analytically outputs a synthetic mesh. 

Our implementation uses the PyTorch framework and the Adam~\cite{kingma2014adam} optimizer, with learning rate 0.001 and batch size 16 (half synthetic and half realistic), running on one GPU until convergence.  We set the weights in our loss function to be $\alpha=0.3$, $\beta=30$, $\gamma=50$ for chairs, $\alpha=0.3$, $\beta=200$, $\gamma=10$ for airplanes, and $\alpha=0.03$, $\beta=4$, $\gamma=1$ for human bodies. (See Sec.~\ref{sec:datasets} for details about datasets.)  All input shapes lie within a sphere of radius 0.6 and are re-centered such that points have zero mean. For all tasks, the actual prediction by the encoder includes the semantic parameters as described above plus a global translation. All datasets are split 4:1 into training and testing sets.

\def\KNNS{\mathit{KNN}\kern-0.15em_S}
\subsection{Deformation Transfer}\label{deform}
To tackle the challenge of detail preservation while performing the editing, we transfer the deformation defined by synthetic shapes to realistic shapes that are semantically close. A new input realistic shape $S_r$ is first encoded into a semantic feature $\tilde{f}_r$. Then, a new feature $\tilde{f}_r'$ is obtained by modifying $\tilde{f}_r$ based on the target editing operation. Both $\tilde{f}_r$ and $\tilde{f}_r'$  are decoded into synthetic shapes. Because the decoder is completely analytical, we can trace the transformation for each point on the synthetic prediction during the editing. Then we can decide on a specific algorithm to define a deformation field which further guides the deformation of the input realistic shape. For example, suppose the per-vertex deformation is $D_s: S_r\mapsto S_r'$, \ie, $v_s'=D_s(v_s), v_s\in S_r,v_s'\in S_r'$. Then we can define the deformation field as:
\[D_r(x)=\frac{\sum_{v\in \KNNS(x,k)}W(x,v)\cdot D_s(v)}{\sum_{v\in \KNNS(x,k)}W(x,v)},\]
where $x\in\mathbb{R}^3$, $\KNNS(x,k)$ is the set of $k$ nearest neighbors on shape $S_r$ of point $x$ ($k=1,2,\dots$), $W(\cdot,\cdot)$ defines the weights among $k$ neighbors and can be either constant or a function of points in the field and on the synthetic shape. In the experiments, we set weights as $W(x,v)=(1+\langle \mathbf{n}_{x},\mathbf{n}_{v}\rangle)^{k_n}\cdot e^{-\frac{\|v-x\|^2}{\sigma^2}} (k_n=2,\sigma=0.03)$
where $\mathbf{n}_p$ is the normal of vertex $p$ ($p=x, v$) for rigid shapes and $W(x,v)=1$ for non-rigid shapes.

\section{Datasets}\label{sec:datasets}

We consider both rigid and non-rigid examples to thoroughly evaluate our proposed method. 

\paragraph{Rigid Shapes}
We consider chairs and airplanes from the ShapeNet~\cite{shapenet2015} dataset, a commonly used large-scale dataset for 3D model evaluation. In order to generate editable synthetic data, we create a template that captures important semantic parameters for each class. Each synthetic chair is composed of 6 cuboids with a predefined structure and eight semantic parameters: the height, depth, and width of the back, seat, and leg.\footnote{Since chair back and seat share the same width in practice, the width of both parts is merged into one parameter.}  The synthetic airplane is a simplified airplane with a smoothed fuselage, two wings, and one vertical and two horizontal stabilizers. We define six semantic parameters: the height, length, and width of the fuselage, and the lengths of the wings, vertical stabilizer, and horizontal stabilizers. The dataset includes 200 and 620 realistic shapes for chairs and airplanes, respectively, and 40000 synthetic shapes for each class. All inputs are point clouds with 2840 and 2750 vertices for chairs and airplanes.

\paragraph{Non-Rigid Shapes}
We conduct experiments on a realistic dataset of human bodies sampled from two sources.  First, we include 5663 3D scans of human bodies from DFAUST~\cite{dfaust:CVPR:2017}.  To incorporate the additional challenge of clothing, we also include 962 clothed bodies from Buff~\cite{Zhang_2017_CVPR}.  To generate a dataset with edits, we consider SMPL~\cite{SMPL:2015}, a synthetic shape template that models pose and shape parameters of the human body. Specifically, the semantic space covers 69 pose parameters (23 joints with 3 degrees of freedom each) and 3 shape parameters, the linear combination of which corresponds to a 6890-vertex synthetic mesh. It is straightforward to deform meshes reconstructed from SMPL~\cite{SMPL:2015} with known shape/pose parameters, which is a much more challenging task for real-world scanned data. The synthetic data is randomly sampled from SMPL parameter space with a Gaussian distribution prior, and all testing shapes are held-out characters not seen during training.

\begin{table}[t]
\caption{\textbf{Dataset and task overview.} Results are presented for 3 shape classes; edits include structure-aware part deformation for man-made objects and human body animation. }
\label{tab:data}
\centering
\addtolength{\tabcolsep}{-4.4pt}
\begin{adjustbox}{width=\linewidth}
\begin{tabular}{cccccc}
\toprule
\textbf{Class}&\textbf{Operation\footnote{Aniso.\ Scale stands for local (per-part) anisotropic scaling, and Rot.\ stands for rotation.}} & \textbf{Synthetic Data} & \textbf{Realistic Data}\\
\midrule
Chair   & Aniso.\ Scale &Cuboids & ShapeNet~\cite{shapenet2015}\\
Airplane  & Aniso.\ Scale & Simplified Airplane & ShapeNet~\cite{shapenet2015}\\
Human & Aniso.\ Scale, Rot. &SMPL~\cite{SMPL:2015} & DFAUST~\cite{dfaust:CVPR:2017},  Buff~\cite{Zhang_2017_CVPR}\\
\bottomrule
\end{tabular}
\end{adjustbox}
\end{table}


\section{Evaluation}\label{sec:experiment}

We test the proposed 3D shape editing system on the tasks of piecewise anisotropic scaling and non-rigid deformation, for the three datasets in Tab.~\ref{tab:data}. Sec.~\ref{sec:rigid} and \ref{sec:nonrigid} qualitatively demonstrate editing of  semantic parameters on rigid and non-rigid datasets, while Sec.~\ref{sec:out-distribution} further tests the method's generalization on out-of-distribution examples.  Sec.~\ref{sec:comparison} shows comparisons to Neural Cages~\cite{yifan2019neural}, a recent technique for source-to-target matching with a learned deformation, and DualSDF~\cite{hao2020dualsdf}, a recent deformation method with learned resynthesis. For each of the above sections, we provide additional results in the Supplementary Material. Finally, Sec.~\ref{sec:ablation} presents ablation studies on key components of the proposed method.  The proposed method requires approximately 3~ms for encoding, and deformation can take tens of milliseconds through seconds, depending on object size.  For comparison, DualSDF requires approximately 175 ms for their gradient descent procedure, and seconds to tens of seconds for ray-tracing results.

\def\imwidth{0.8\linewidth}
\begin{figure*}[t]
 \centering
 \includegraphics[width=\imwidth]{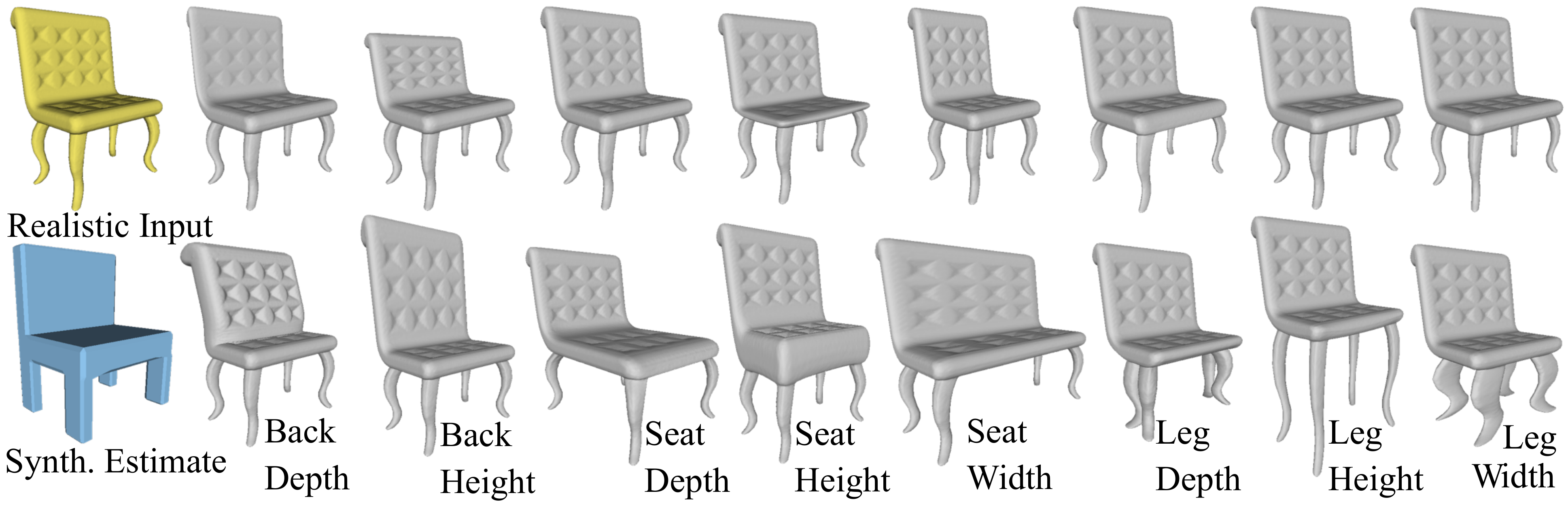}
 \includegraphics[width=\imwidth]{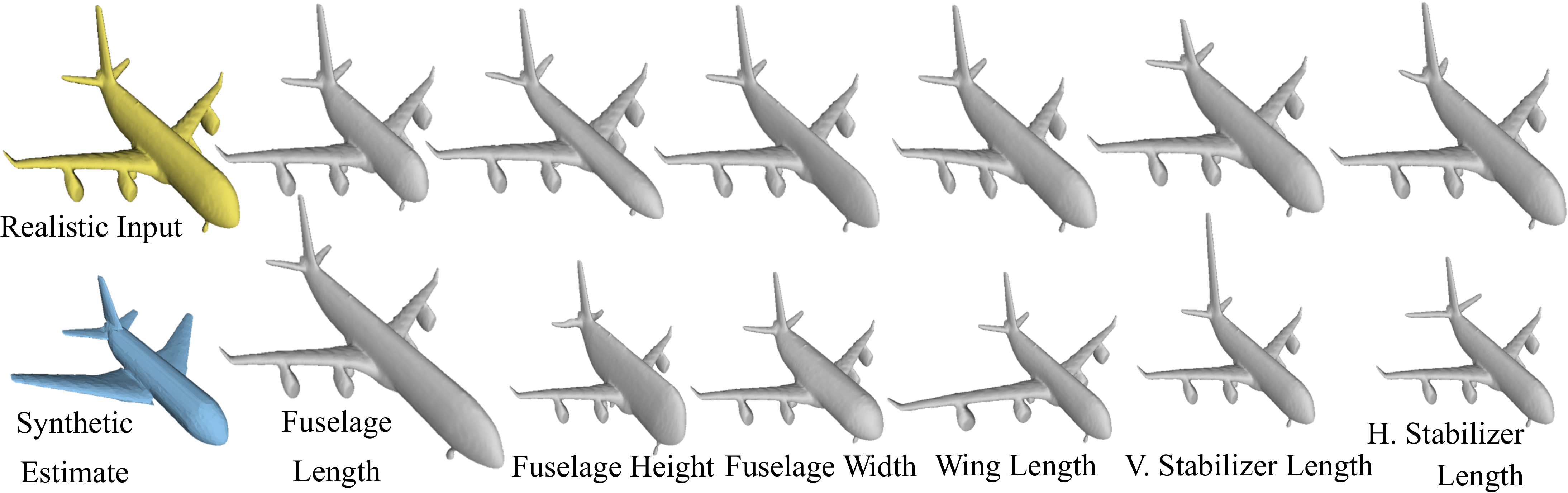}
 \caption{\textbf{Editing results with piecewise anisotropic scaling.} Realistic input shapes (yellow) are fit to synthetic templates (blue), then edited by decreasing (top row) and increasing (bottom row) each shape parameter in turn.  The proposed method provides for both detail preservation and independent control over each semantic parameter.}
 \label{fig:sem-edit-rigid}
\end{figure*}

\begin{figure*}
 \centering
 \includegraphics[width=\imwidth]{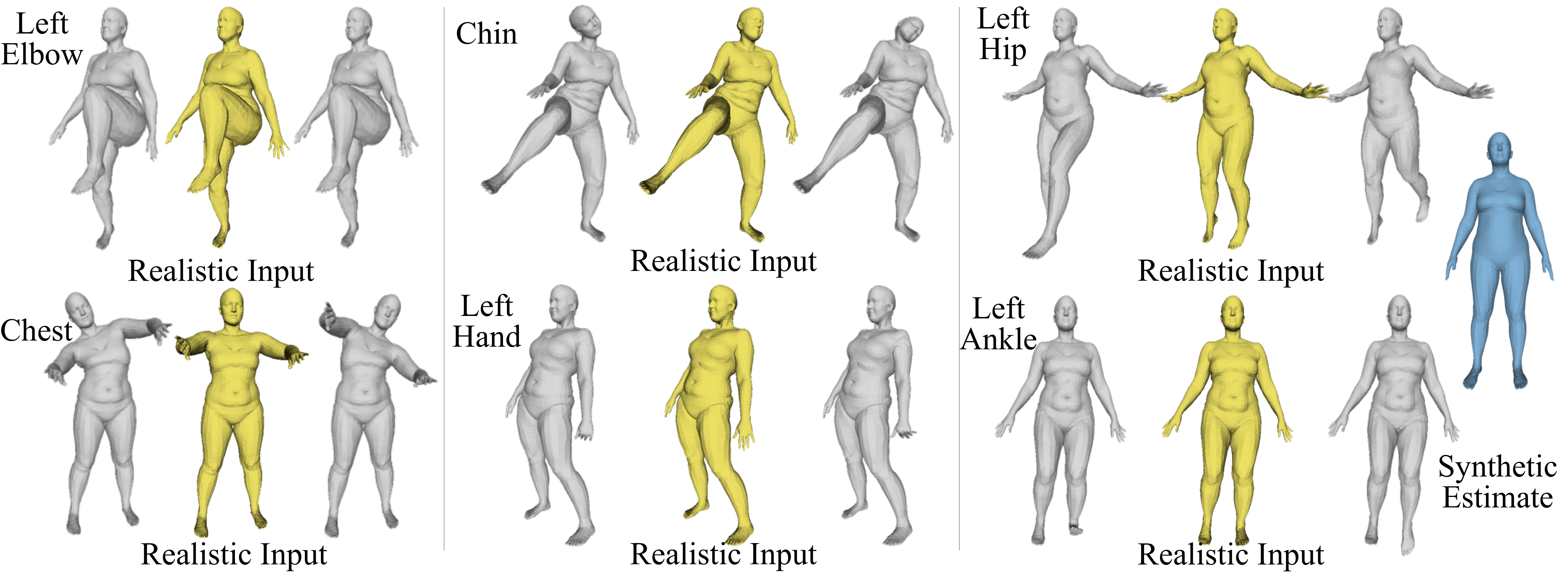}
 \caption{\textbf{Editing results for deformation of human bodies.} In each set of 3 shapes, an input human body (yellow) is deformed by changing one semantic parameter in both directions. An example synthetic template (blue) is shown at right. }
 \label{fig:sem-edit-nonrigid}
\end{figure*}

\subsection{Piecewise Anisotropic Scaling Results}\label{sec:rigid}
A common editing strategy for objects with well-defined semantic parts is anisotropic scaling of different parts independently.
In Fig.~\ref{fig:sem-edit-rigid}, we show edits produced by the proposed method for chairs and airplanes (which were held out from training). For each class, we show an input shape (yellow) and the estimated synthetic template (blue). Note that the latter matches the realistic shape in all structural components relevant to editing, such as the seat of the chair and wing length of the airplane. This shows that the model correctly infers semantic parameters of realistic shapes, though trained with only synthetic parameter annotations. In gray, we show the deformation result of one semantic parameter per column, with the first row decreasing and the second row increasing that parameter. As hoped, modifying one semantic parameter does not affect the shape of other parts; \eg, modifying the depth of a chair's seat does not change the shape of its legs.
At the same time, parameters correctly update groups of similar parts. For example, modifying the ``leg height'' parameter for chairs updates all four legs, while modifying ``wing length'' updates both airplane wings. Note that details are well-preserved after deformation (\eg, tufts on the chair back and seat, curved chair legs, engines and landing gear of the airplane). The proposed method allows for substantial parameter change (\eg, height of chair back and legs, width of chair seat, length of airplane fuselage), in contrast to methods that train on shape databases that might lack such extreme deformations.

\subsection{Non-Rigid Deformation Results}\label{sec:nonrigid}
In Fig.~\ref{fig:sem-edit-nonrigid}, we show results for editing semantic pose parameters of human bodies. Our system is able to apply the deformations defined by a simple, parameterized skinned model (SMPL) to realistic human input shapes (DFAUST), accommodating both large (hip movement) and subtle (ankle movement) motions. Note that each joint has up to three degrees of freedom, though because of space constraints we demonstrate only one degree of freedom per joint.  Our model correctly decouples the effects of all parameters, even those affecting the same joint. 

\begin{figure}[t]
   \includegraphics[width=\linewidth]{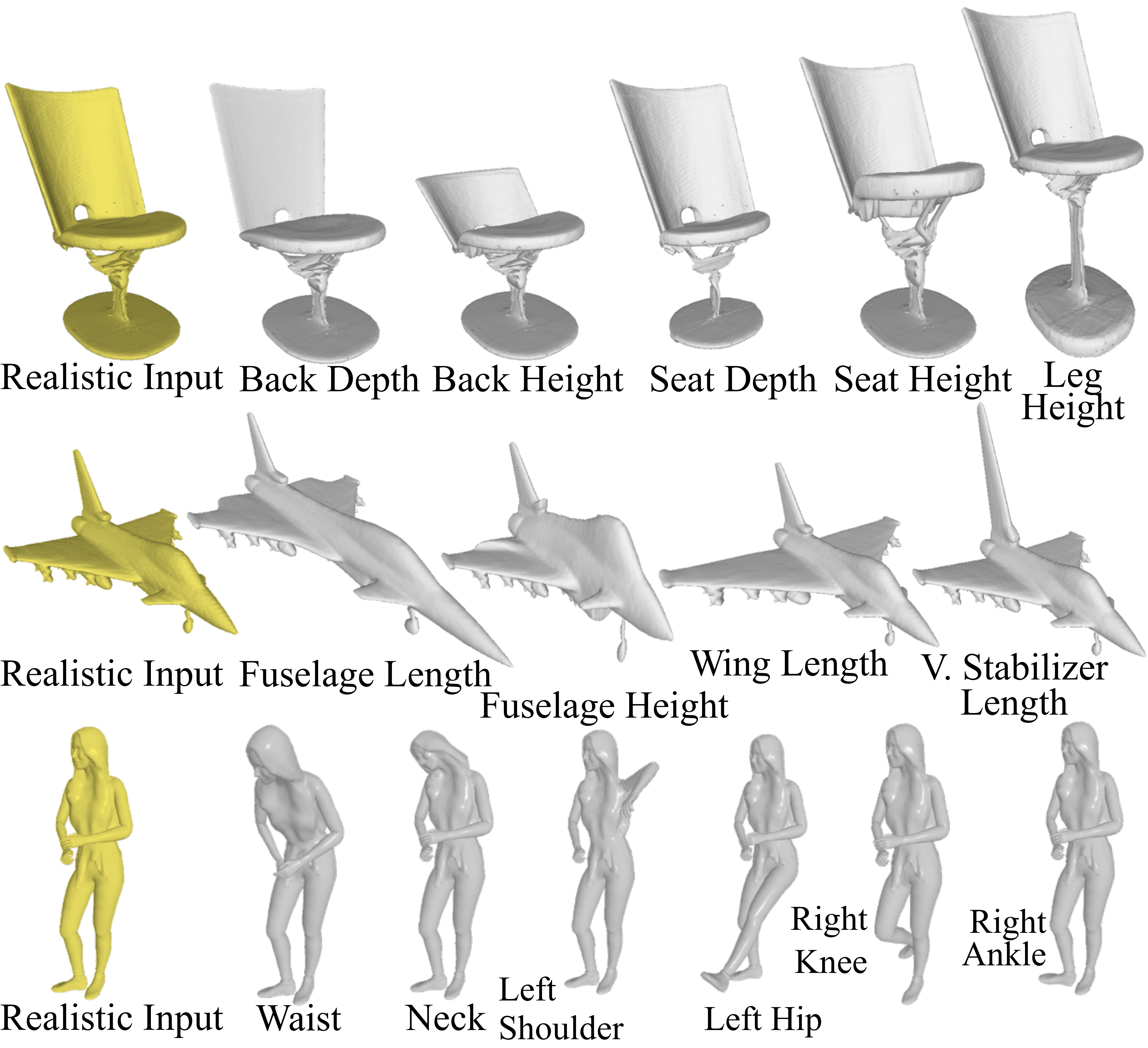}
   \caption{\textbf{Editing results for out-of-distribution shapes.}  Our method produces reasonable results, and correctly preserves detail, even if the input shape has missing parts or different topology relative to the synthetic template.}
\label{fig:out-distribution}
\end{figure}

\subsection{Out-of-Distribution Shapes}\label{sec:out-distribution}
To evaluate the generalization ability of the proposed method, we test it on shapes falling outside of the shape distribution of the training dataset.  Fig.~\ref{fig:out-distribution} shows examples from all three classes, illustrating cases in which the test shapes are topologically different from the training examples or are missing some of the components present in the synthetic shapes.  For example, the chair at the top has only one leg, the airplane has a delta wing and is missing horizontal stabilizers, while the human has long hair and an initial pose with the hands almost close together.  In all cases, the proposed method produces reasonable deformed results and preserves input shape details. 

\subsection{Comparison to Prior Work}\label{sec:comparison}

\def\imwidth{0.86\linewidth}
\begin{figure}[t]
   \centering
   \includegraphics[width=\imwidth]{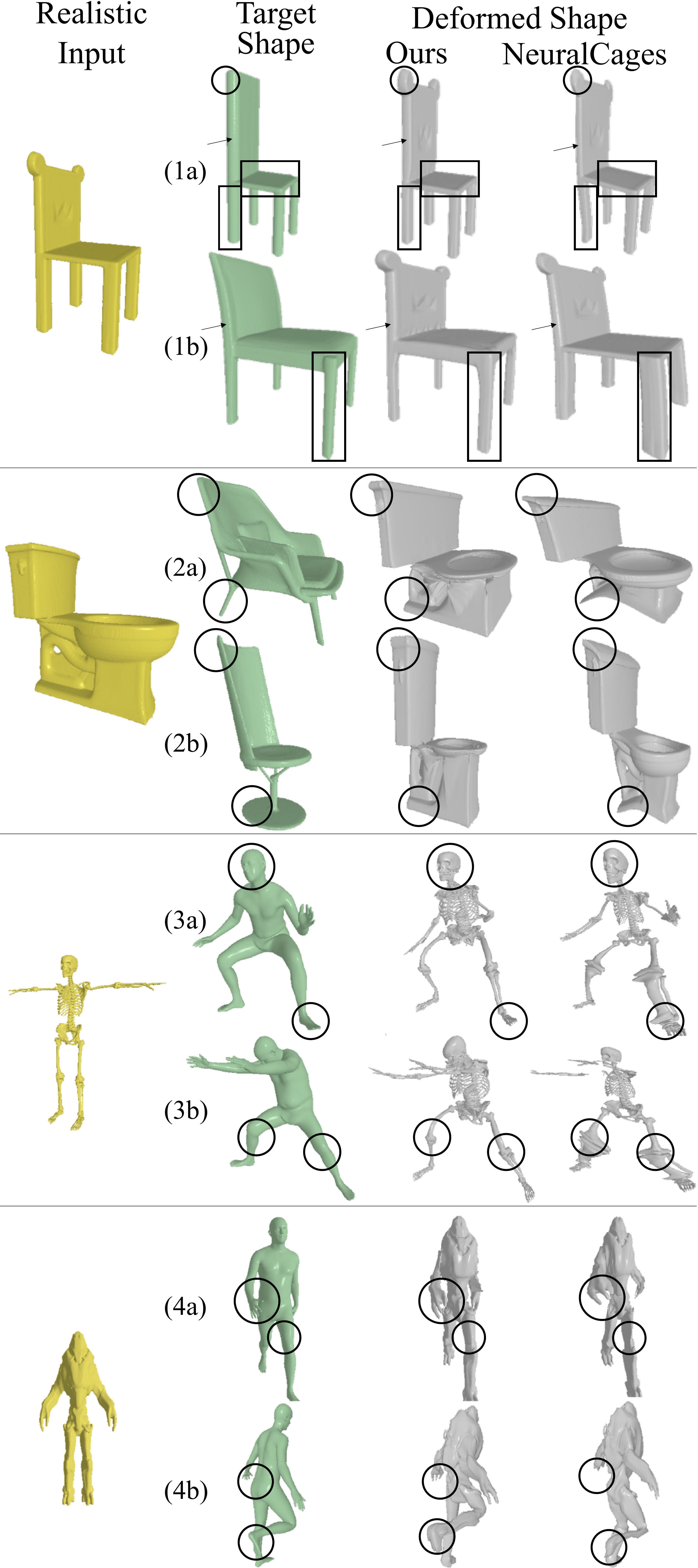}
\caption{\textbf{Comparison to Neural Cages~\cite{yifan2019neural}.} Each source shape (yellow) is deformed to match two target shapes (green), using both the proposed method and Neural Cages. Both are able to globally match the target shape, but Neural Cages often exhibits distortion in regions of large deformation, and an inability to match semantic parameters if local deformation is required.  Our results support more accurate and granular manipulation, even for extreme poses.
 }
  \vspace{-5pt}
\label{fig:comp-neuralcage}
\end{figure}

\begin{figure}[t!]
   \includegraphics[width=1\linewidth,trim={0cm 0cm 0cm 0cm},clip]{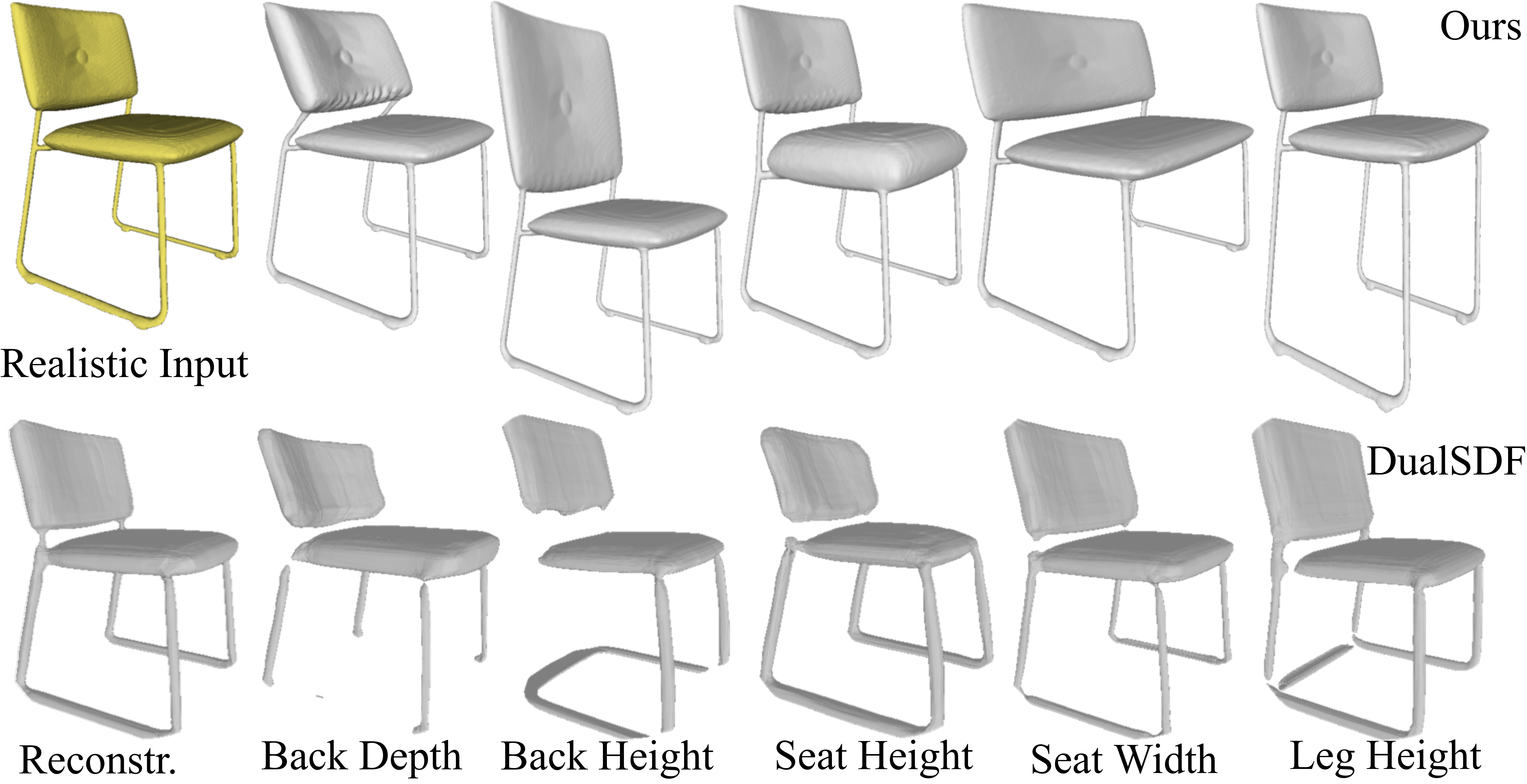}
   \caption{\textbf{Comparison to DualSDF~\cite{hao2020dualsdf}.} The input (yellow) is edited by changing a semantic parameter in our system, or adjusting the radius of a primitive in DualSDF.  When changing a local parameter with DualSDF, the global shape is also affected, as seen in the alteration or omission of the bottom crossbars on the legs. In addition, in contrast to the proposed method, the DualSDF results lack details such as the button on the chair back.}
\label{fig:comp-dualsdf}
\end{figure}

To compare against Neural Cages~\cite{yifan2019neural}, which was designed for the task of source-to-target deformation rather than direct manipulation of semantic parameters, we modify the proposed method to use a deformation determined by templates fit to both the source and target shapes. Fig.~\ref{fig:comp-neuralcage} shows four examples, including both in- and out-of-distribution chairs and two human body shapes evaluated in the original Neural Cages work.
Both methods are able to match the target shape globally. However, note the spurious deformation in the legs in the Neural Cages results for (1a) and (1b), as well as the fact that in (1b) the Neural Cages result matches the seat thickness of the source, not the target.  In contrast, the proposed method does not introduce unnecessary deformations, yet is able to deform the source locally, not just globally, to match the seat thickness. In the human body results, Neural Cages can introduce distortion in regions with large deformation, such as the knees in example (3b) or disproportionately scaled hands in examples (4a) and (4b). We observe that, compared to deformation methods based on coarse geometric handles, our results support more granular manipulation even with extreme poses.


We also compare the proposed method with DualSDF~\cite{hao2020dualsdf} on the task of part manipulation. We consider examples common to DualSDF's training set and our testing set. To generate results for DualSDF, we first obtain the embedding of an input shape and then manually adjust the radius of a primitive on the corresponding part along the corresponding dimension. Fig.~\ref{fig:comp-dualsdf} shows the results for one chair example.  The DualSDF reconstruction, which requires re-synthesis, loses details of the original shape such as the button on the back and the four anti-skid pads on the legs (please zoom in for a clear view). In addition, manipulating a single primitive in one part will cause deformation in other parts as well, since the latent embeddings from DualSDF are highly entangled. For example, when manipulating primitives on the back, the leg topology is also changed by DualSDF, but correctly preserved by the proposed method. 

\subsection{Ablation Studies}\label{sec:ablation}
We analyze the influence of model losses on the accuracy of edits to synthetic shapes (chairs).  As shown in Tab.~\ref{tab:ablation-1}, the semantic parameter loss ($L_{sem}$) alone is more effective than shape reconstruction loss ($L_{rec}$) alone. Combining both loss terms achieves the best performance on synthetic shapes than either single loss. From the last two rows, it can be seen that the editing branch for synthetic shapes is essential for reducing mean vertex error (MVE). Also, we observe that models with an editing branch are more stable during training.  Finally, we have evaluated the effects of using either $\ell_1$ or $\ell_2$ distance for $L_{sem}$; we find that $\ell_2$ performs better in our training.

 \begin{table}[h!]
  \caption{\textbf{Effectiveness of loss terms and the editing branch on the proposed model.} We show the percentage of vertices in synthetic testing shapes (from the chair class) having Mean Vertex Error (MVE) below a given threshold. For example, the rightmost entry of the bottom row indicates that 95.4\% of vertices are within a distance of 0.03 of their ground-truth locations.}
 \label{tab:ablation-1}
 \addtolength{\tabcolsep}{-3pt}
 \begin{adjustbox}{width=\linewidth}
 \begin{tabular}{ccccc}
 \toprule
 \textbf{Loss Terms} & \textbf{Editing Branch} & \textbf{MVE\textless 0.01} & \textbf{MVE\textless 0.02} & \textbf{MVE\textless 0.03} \\
 \midrule
  $L_{sem}$           & \checkmark & 35.2\% & 72.3\% & 87.2\% \\
  $L_{rec}$           & \checkmark & 11.9\% & 47.9\% & 77.6\% \\
  $L_{sem} + L_{rec}$ &            & 11.5\% & 55.0\% & 83.6\% \\
  $L_{sem} + L_{rec}$ & \checkmark & 40.3\% & 82.2\% & 95.4\% \\
 \bottomrule
 \end{tabular}
 \end{adjustbox}
 \end{table}
 
\section{Conclusion and Future Work}
In this paper, we present a learning framework for detail-preserving semantic 3D shape editing. We propose to infer semantic parameters of input examples by leveraging a simple synthetic shape set and learning a joint low-dimensional embedding for synthetic and realistic shapes. This approach allows the proposed method to relieve the need for semantic part labels or example user edits on realistic shapes, while allowing semantically consistent edits for all shapes (including out-of-distribution examples). Experiments on both rigid and non-rigid shapes demonstrate that the proposed method provides detail-preserving, structure-aware semantic editing and compares favorably with prior work. 

This work is a first step toward learning a semantic 3D shape editing system, and there are several ways the proposed method can be extended. Currently, the semantic encoding is learned and the deformation is analytic. Including both in an end-to-end learning pipeline is a valuable future direction. Also, it is interesting to consider how our semantic encoding can be combined with other deformation transfer strategies, possibly involving training on unlabeled shape datasets, to achieve new types of shape edits.

\textbf{Acknowledgement}
We would like to thank Kyle Genova for sharing watertight ShapeNet meshes, and Francesco Locatello, Chiyu ``Max'' Jiang and Jingwei Huang for helpful discussions. We also thank the National Science Foundation (grant \#IIS-1815070) for
partial support of this work.

{\small
\bibliographystyle{ieee}
\bibliography{egbib}
}
\newpage
\appendix
\section{Supplementary Material}
In this Supplementary Material, we provide additional results for the experiments on piecewise anisotropic scaling and non-rigid deformation on the chair, airplane, and human body datasets described in Sec.~\ref{sec:datasets}. Sec.~\ref{sec:rigid} and \ref{sec:nonrigid} demonstrate qualitative semantic parameter editing results on rigid and non-rigid shape categories, respectively, while Sec.~\ref{sec:out-distribution} shows results on out-of-distribution examples.  Sec.~\ref{sec:comparison} provides additional comparisons to Neural Cages~\cite{yifan2019neural}, a recent technique for source-to-target matching with a learned deformation, and DualSDF~\cite{hao2020dualsdf}, a recent deformation method with learned re-synthesis.

\def\imwidth{1\linewidth}
\begin{figure*}[t]
 \centering
 \includegraphics[width=0.86\imwidth]{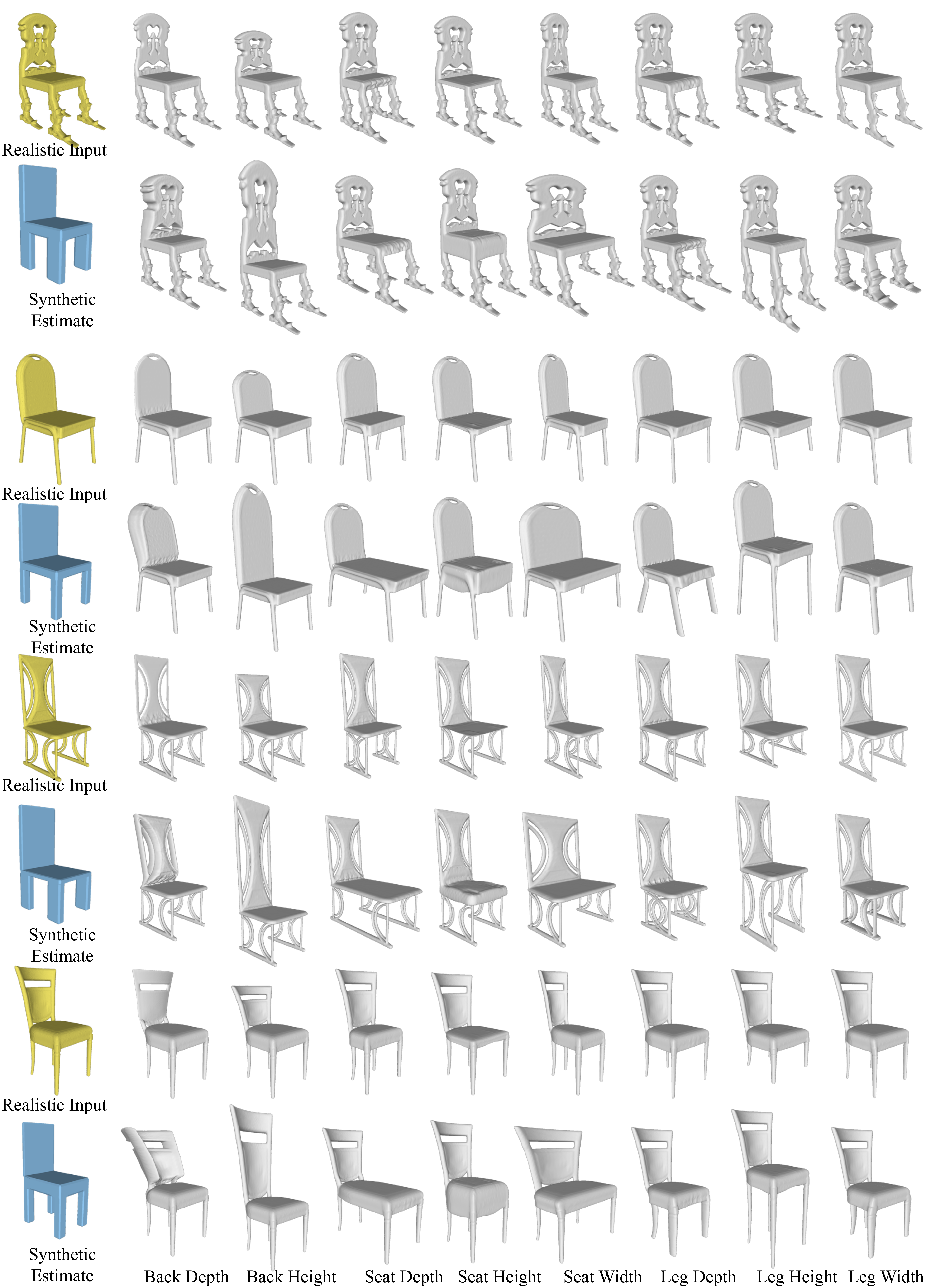}
 \caption{\textbf{Editing results for piecewise anisotropic scaling on chairs.} Realistic input shapes (yellow) are fit to synthetic templates (blue), then edited by decreasing (top row) and increasing (bottom row) each shape parameter. The proposed method preserves details and independently controls each semantic parameter.}
 \label{fig:sem-edit-rigid-chair1}
\end{figure*}
\begin{figure*}[t]
 \centering
 \includegraphics[width=0.86\imwidth]{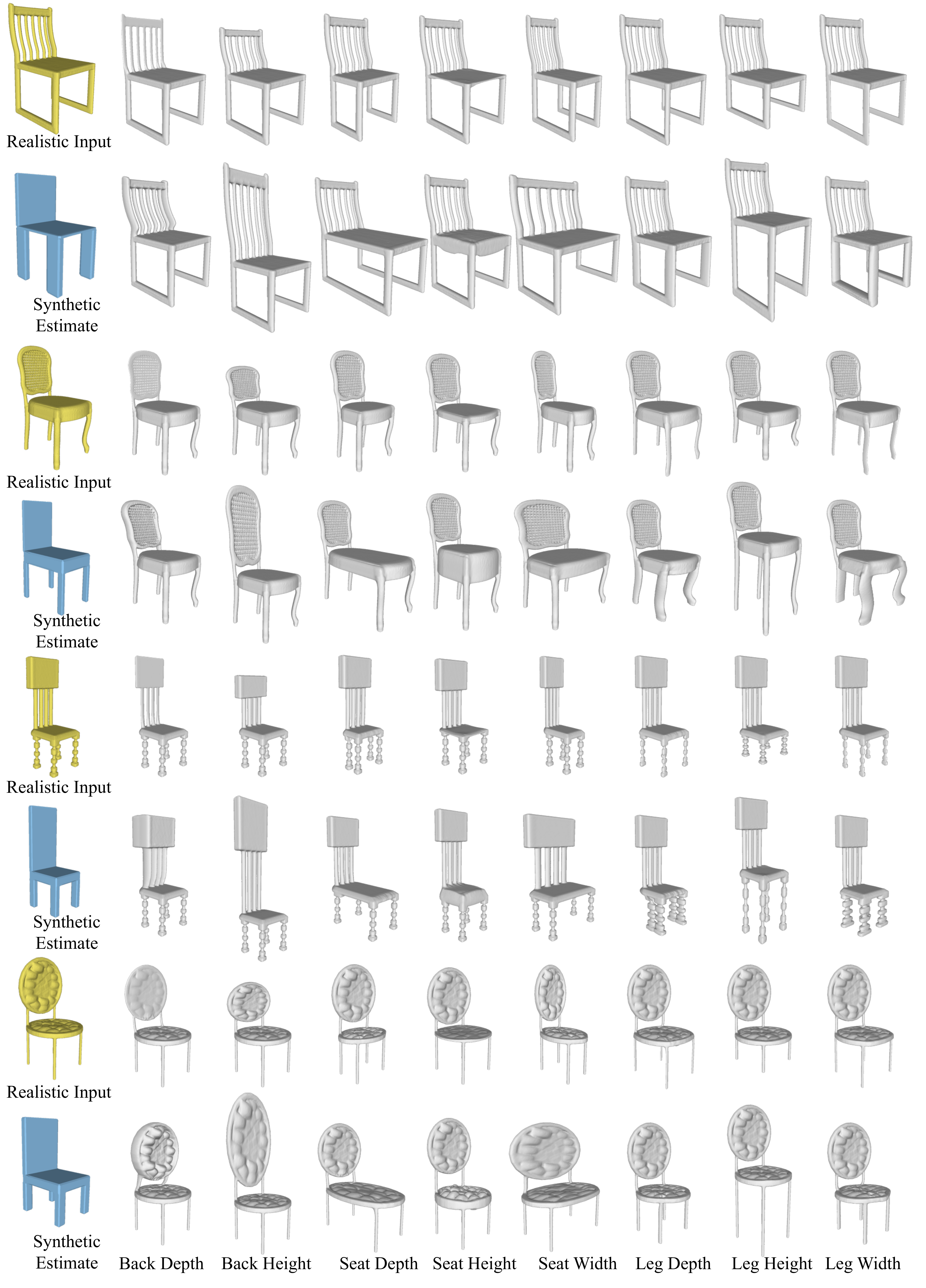}
 \caption{\textbf{Additional editing results with piecewise anisotropic scaling on chairs.} Realistic input shapes (yellow) are fit to synthetic templates (blue), then edited by decreasing (top row) and increasing (bottom row) each shape parameter. The proposed method preserves details and independently controls each semantic parameter.}
 \label{fig:sem-edit-rigid-chair2}
\end{figure*}

\def\imwidth{0.95\linewidth}
\begin{figure*}[t]
 \vspace{-0.2cm}
 \centering
 \includegraphics[width=\imwidth]{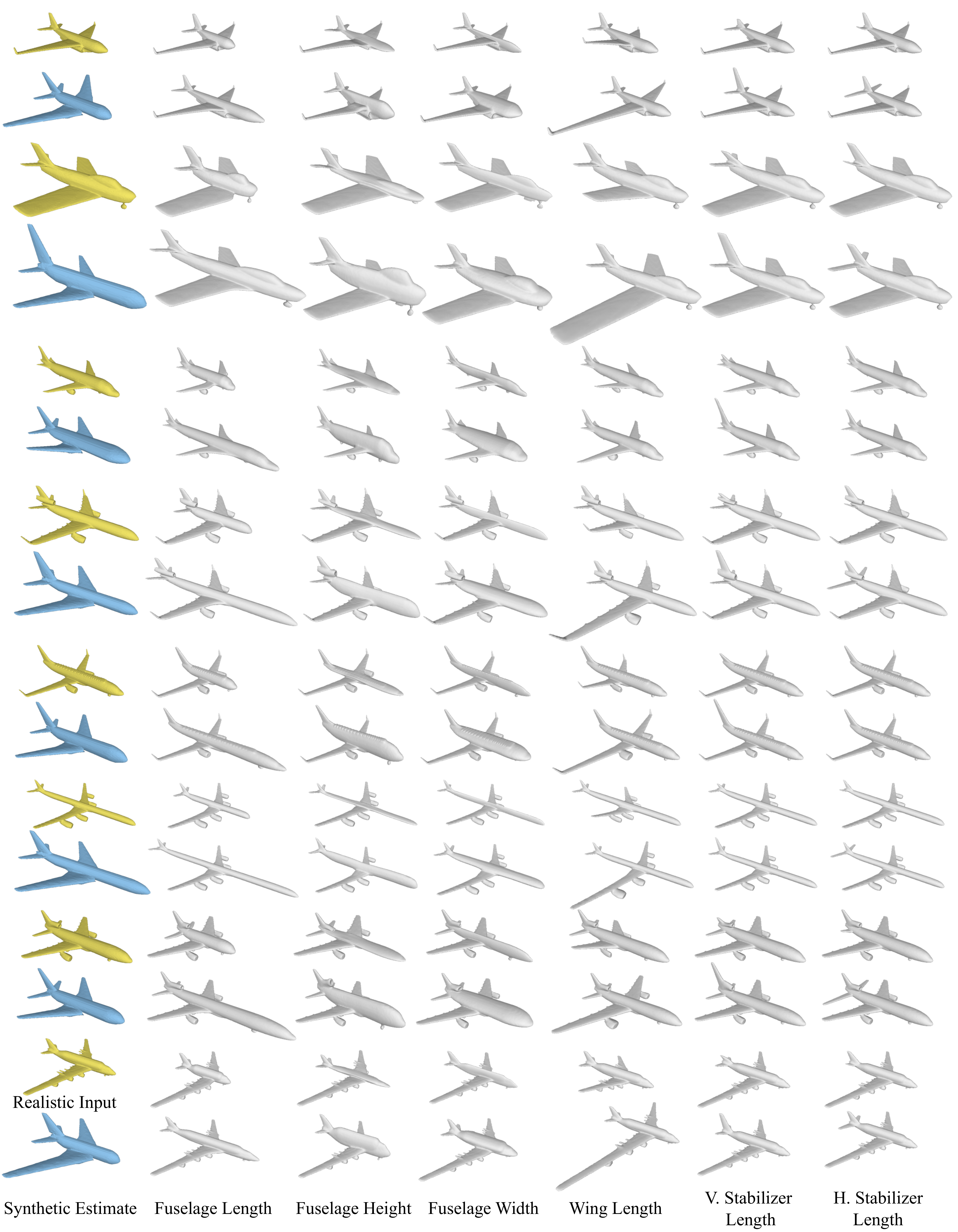}
 \caption{\textbf{Editing results with piecewise anisotropic scaling on airplanes.} Realistic input shapes (yellow) are fit to synthetic templates (blue), then edited by decreasing (top row) and increasing (bottom row) each shape parameter in turn. The proposed method preserves details and independently controls each semantic parameter.}
 \label{fig:sem-edit-rigid-airplane}
\end{figure*}

\begin{figure*}
 \centering
  \includegraphics[width=\imwidth]{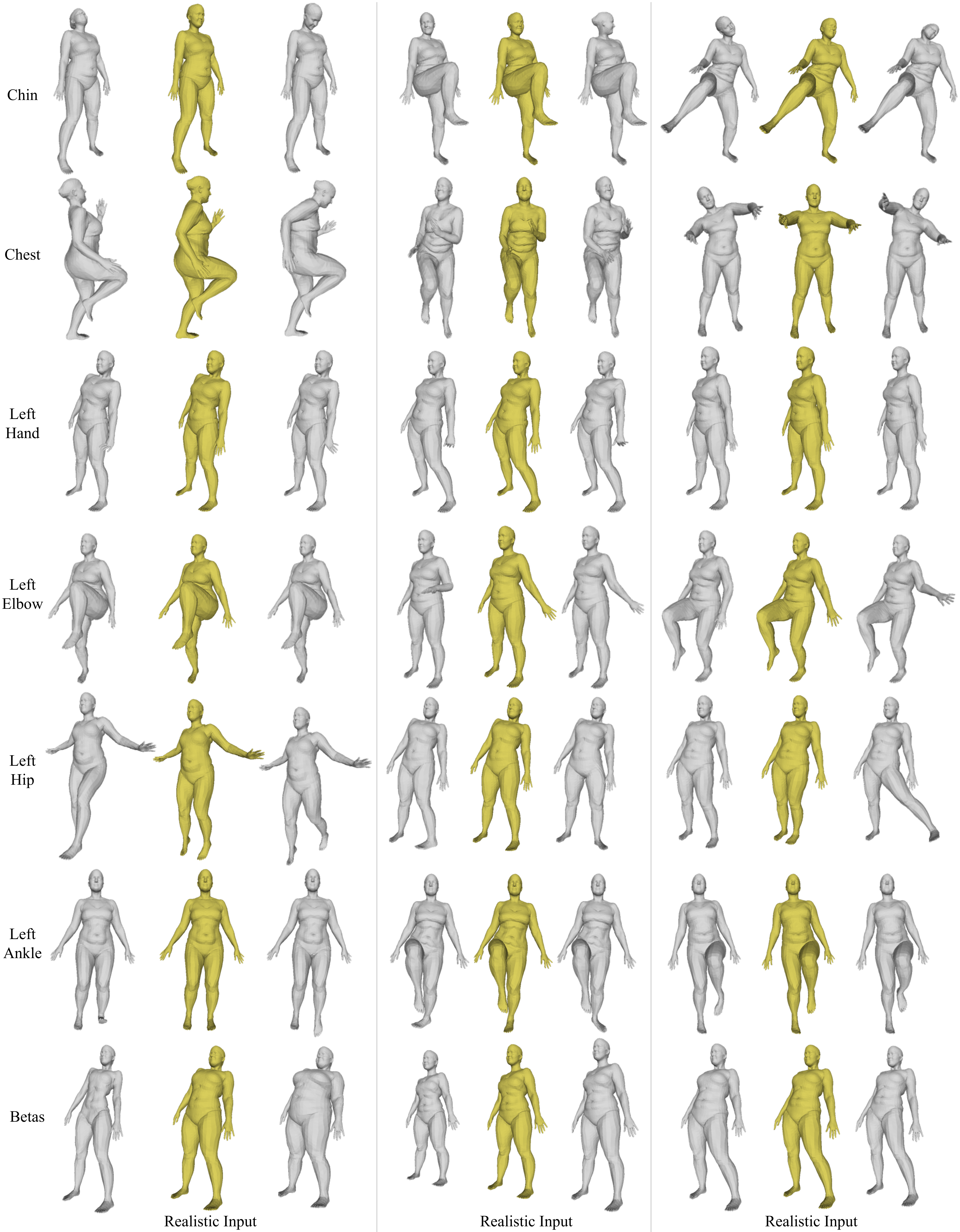}
 \caption{\textbf{Editing results for deformation on human bodies from the DFAUST~\cite{dfaust:CVPR:2017} dataset}. In each set of three shapes, an input human body (yellow) is deformed by changing one semantic parameter in both directions. The first six rows show results of deforming three degrees of freedom of all joints presented in Fig.~\ref{fig:sem-edit-nonrigid}, with one joint in each row. The last row shows deformation results of the three shape parameters (betas). }
 \label{fig:sup-sem-edit-nonrigid}
\end{figure*}

\begin{figure*}
 \centering
 \includegraphics[width=\imwidth]{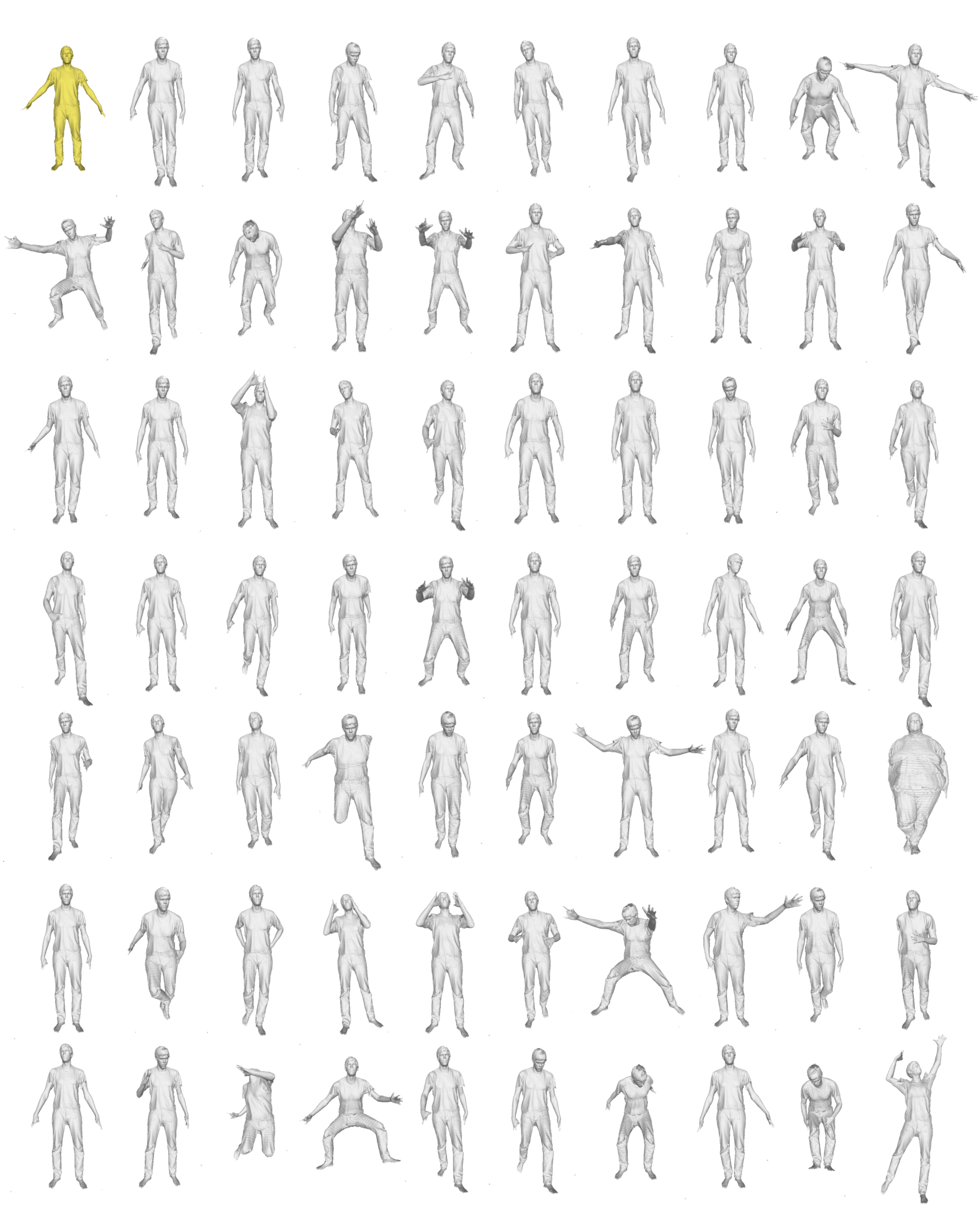}
 \caption{\textbf{Editing results for deformation of human bodies from the Buff~\cite{Zhang_2017_CVPR} dataset.} Given an input testing shape, we are able to arbitrarily manipulate any set of semantic parameters and output the deformed shapes.}
 \label{fig:sup-sem-edit-nonrigid2}
\end{figure*}

\subsection{Piecewise Anisotropic Scaling Results}\label{sec:rigid}

A common editing strategy for objects with well-defined semantic parts is anisotropic scaling of different or independent parts.
We show anisotropic editing results produced by the proposed method for chairs in Fig.~\ref{fig:sem-edit-rigid-chair1}   and Fig.~\ref{fig:sem-edit-rigid-chair2}, and for airplanes in 
Fig.~\ref{fig:sem-edit-rigid-airplane}. All shown examples are testing examples, \ie, held out during the training stage. For each class, we show an input shape (yellow) and the estimated synthetic template (blue). Note that the latter matches the realistic shape in all structural components relevant to editing, such as the seat of the chair or wing length of the airplane. Therefore, the proposed model correctly infers semantic parameters of realistic shapes when trained with only synthetic parameter annotations. In gray, we show the deformation result of one semantic parameter per column, with the first row decreasing and the second row increasing that parameter. As hoped, modifying one semantic parameter does not affect other parameters, \ie, the shape of other parts. For example, modifying the depth of a chair seat does not change the shape of its legs.
At the same time, parameters correctly update groups of similar parts. For example, modifying the ``leg height'' parameter for a chair updates all four legs, while modifying ``wing length'' updates both airplane wings. Note that details are well-preserved after deformation (\eg, patterns on the chair back and seat, curved chair legs, engines and landing gear of the airplane). The proposed method allows for substantial parameter change (\eg, height of chair back and legs, width of chair seat, length of airplane fuselage), in contrast to methods that train on shape databases supervised with ground truth deformed shapes that might lack such extreme deformations.

One advantage of the proposed method is to support continuous deformation. Please visit our project page for such results.

\subsection{Non-Rigid Deformation Results}\label{sec:nonrigid}
In Fig.~\ref{fig:sup-sem-edit-nonrigid} and Fig.~\ref{fig:sup-sem-edit-nonrigid2}, we show results for editing semantic pose parameters of human bodies on DFAUST~\cite{dfaust:CVPR:2017} and Buff~\cite{Zhang_2017_CVPR} datasets, respectively. Both source inputs are characters held out during the training stage. Our system is able to apply the deformations defined by a simple, parameterized skinned model (SMPL~\cite{SMPL:2015}) to realistic human input shapes, accommodating both large (hip movement) and subtle (ankle movement) motions. Note that in Fig.~\ref{fig:sem-edit-nonrigid}, we demonstrate only one degree of freedom per joint because of space constraints. In Fig.~\ref{fig:sup-sem-edit-nonrigid}, we provide full deformation results for all three degrees of freedom of each joint presented in Fig.~\ref{fig:sem-edit-nonrigid}. We can see that the proposed method accepts and recognizes the parameters of shapes in various poses. Our model also correctly decouples the effects of all parameters, even those affecting the same joint. In Fig.~\ref{fig:sup-sem-edit-nonrigid2}, we deform an arbitrary set of shape and pose parameters for the same input shape (yellow). Note that this input shape is a scanned clothed male and is noisy, not watertight, and with many isolated faces around the body.


\begin{figure*}[b]
   \centering
  \includegraphics[width=\linewidth]{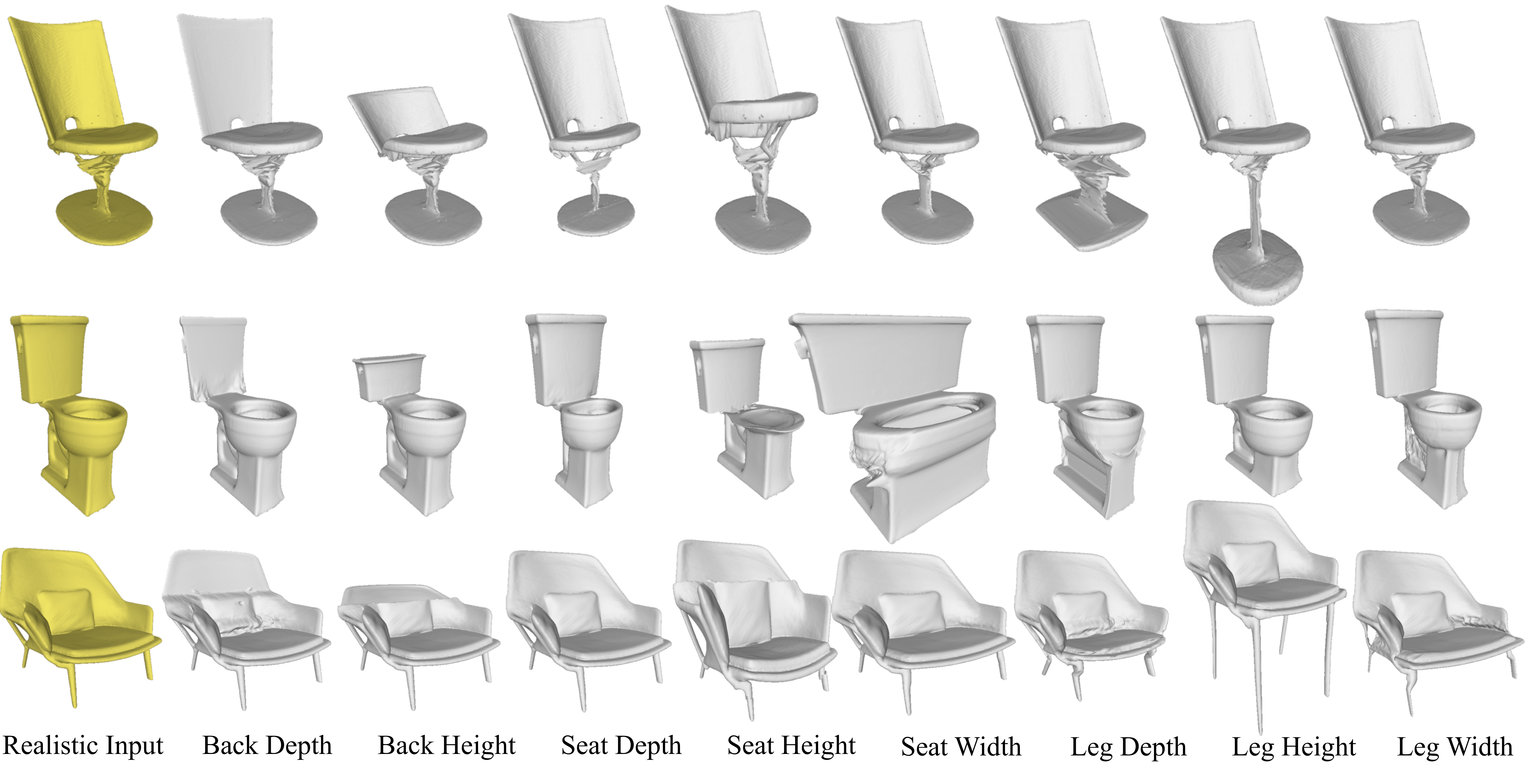}
   \includegraphics[width=\linewidth]{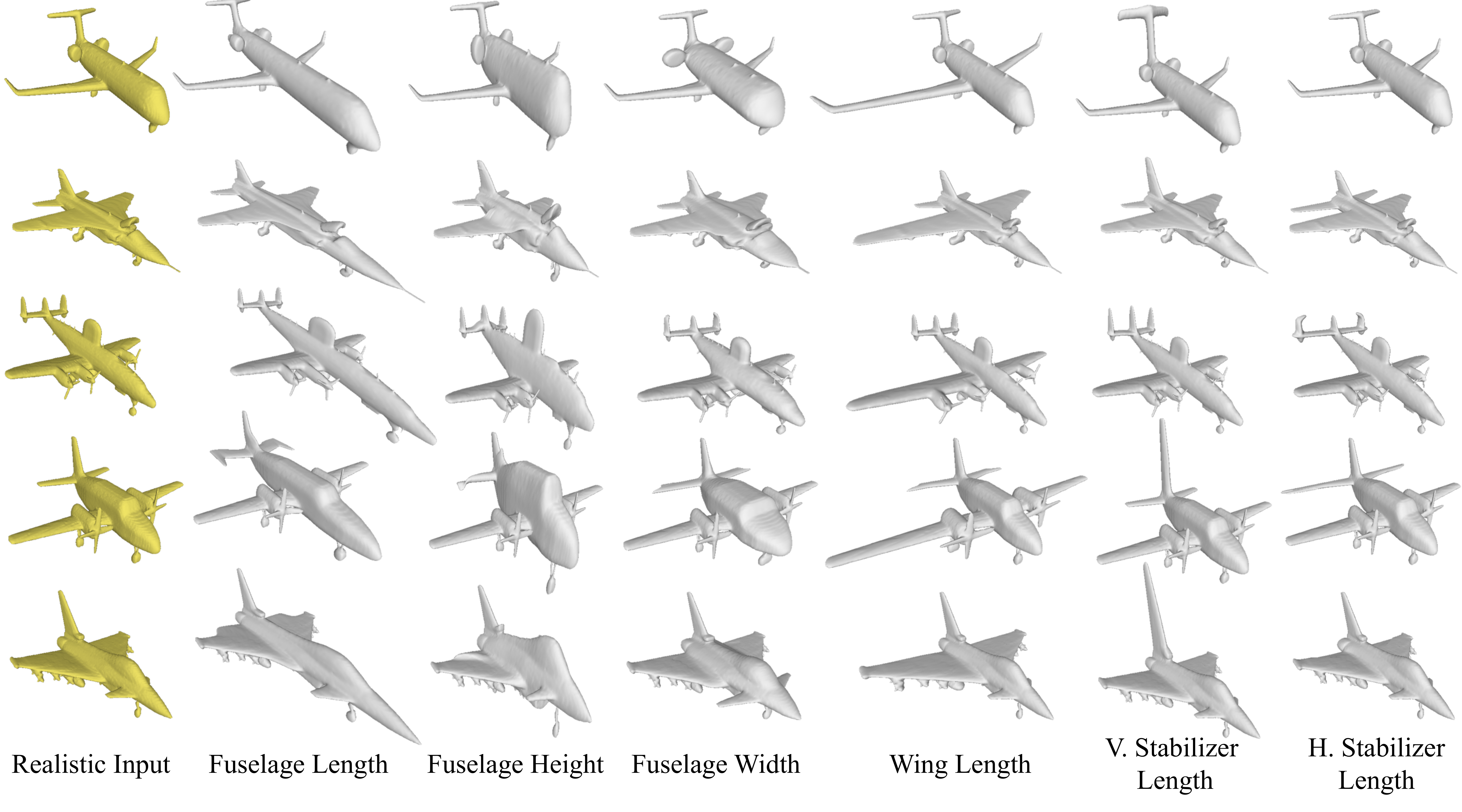}
   \caption{\textbf{Editing results for out-of-distribution shapes on chairs and airplanes.}  Our method produces semantically meaningful results, and correctly preserves detail, even if the input shape has missing parts or different topology relative to the synthetic template.}
\label{fig:out-distribution1}
\end{figure*}
\begin{figure*}[t]
   \centering
  \includegraphics[width=\linewidth]{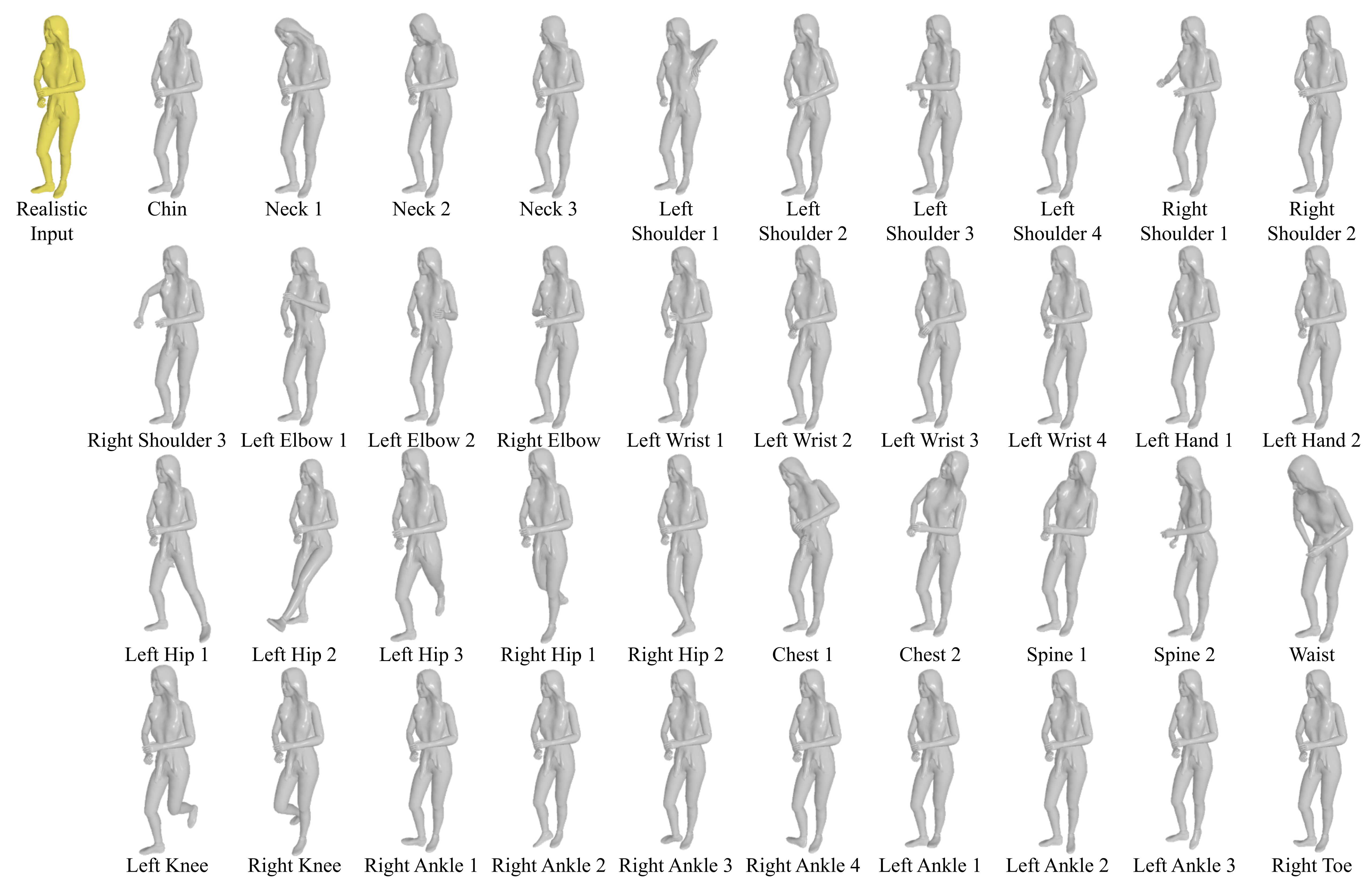}
   \caption{\textbf{Editing results for out-of-distribution shapes on human bodies.}  Our method produces semantically meaningful results, and correctly preserves detail, even if the input shape has missing parts or topology different from the synthetic template.}
\label{fig:out-distribution2}
\end{figure*}

\subsection{Out-of-Distribution Shapes}\label{sec:out-distribution}
To evaluate the generalization ability of the proposed method, we test it on shapes falling outside of the shape distribution of the training dataset.  Fig.~\ref{fig:out-distribution1} and Fig.~\ref{fig:out-distribution2} show examples from all three classes, illustrating cases in which the test shapes are topologically different from the training examples or are missing some of the components present in the synthetic shapes.  For example, in  Fig.~\ref{fig:out-distribution1}, the first chair (top row) and the toilet in the second row have only one leg each, the chair in the bottom row has arms and pillows. The first airplane has horizontal stabilizers on the top, the third airplane has three vertical stabilizers, the second and last airplanes have a delta wing, the third and fourth airplanes have straight wings and propellers on the wings, while in  Fig.~\ref{fig:out-distribution2} the human has long hair and an initial pose with the hands almost close together. Note that each joint has up to three degrees of freedom and can be rotated in two directions for each degree of freedom, therefore there are many deformation results under the same parameter name in Fig.~\ref{fig:out-distribution2}. In all cases, the proposed method produces semantically meaningful results and preserves input shape details.

\subsection{Comparison to Prior Work}\label{sec:comparison}

\def\imwidth{1\linewidth}
\begin{figure*}[t]
   \centering
   \includegraphics[width=0.95\imwidth]{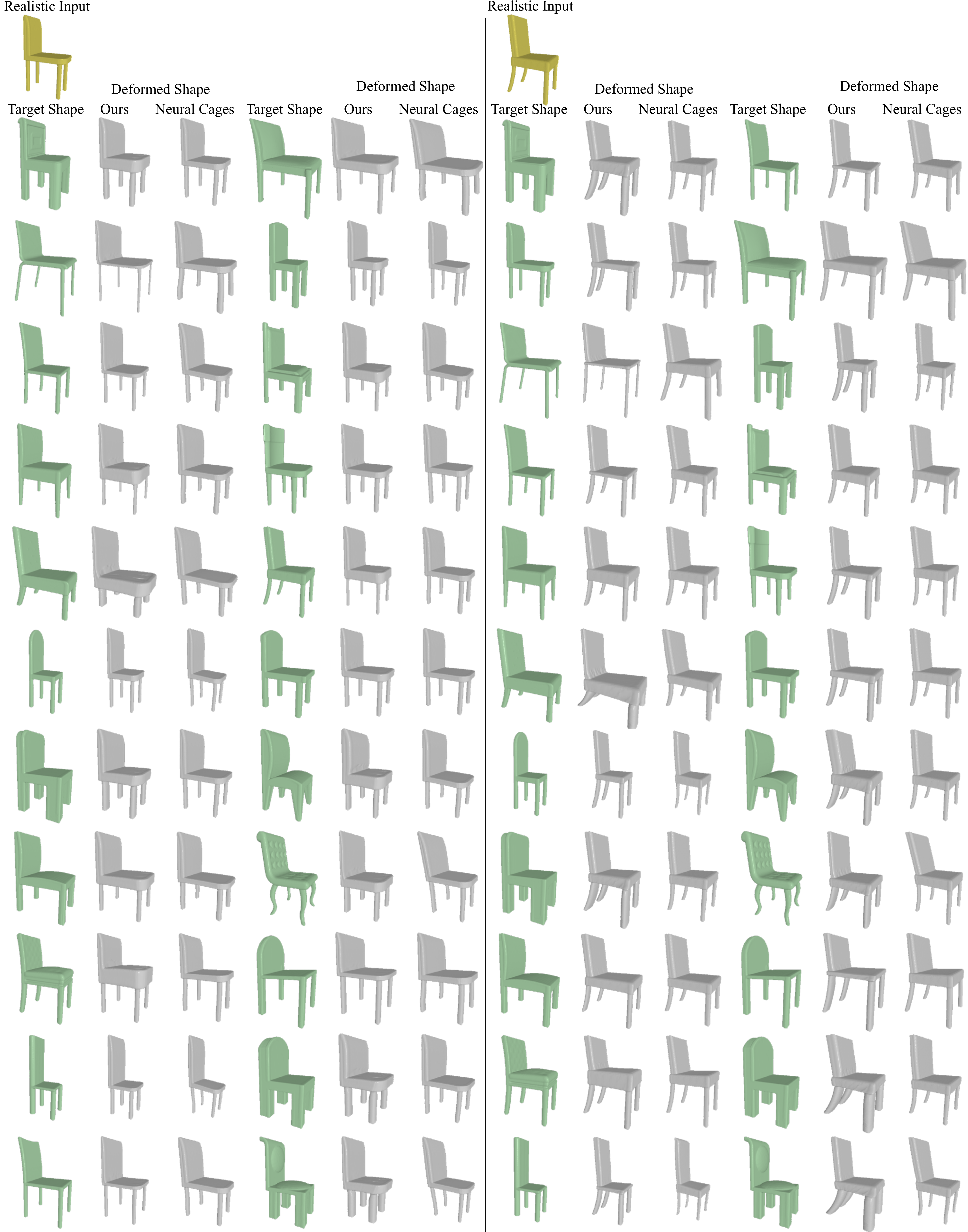}
\caption{\textbf{Comparison to Neural Cages~\cite{yifan2019neural} (chairs).} Each source shape (yellow) is deformed to match 22 target shapes (green), using both the proposed method and Neural Cages. Both methods are able to globally match the target shape, but Neural Cages often exhibits distortion in regions of large deformation and cannot match semantic parameters if local deformation is required. In contrast, our method support more accurate and granular manipulation.}
\label{fig:comp-neuralcage-chair1}
\end{figure*}
\begin{figure*}[t]
   \centering
   \includegraphics[width=0.95\imwidth]{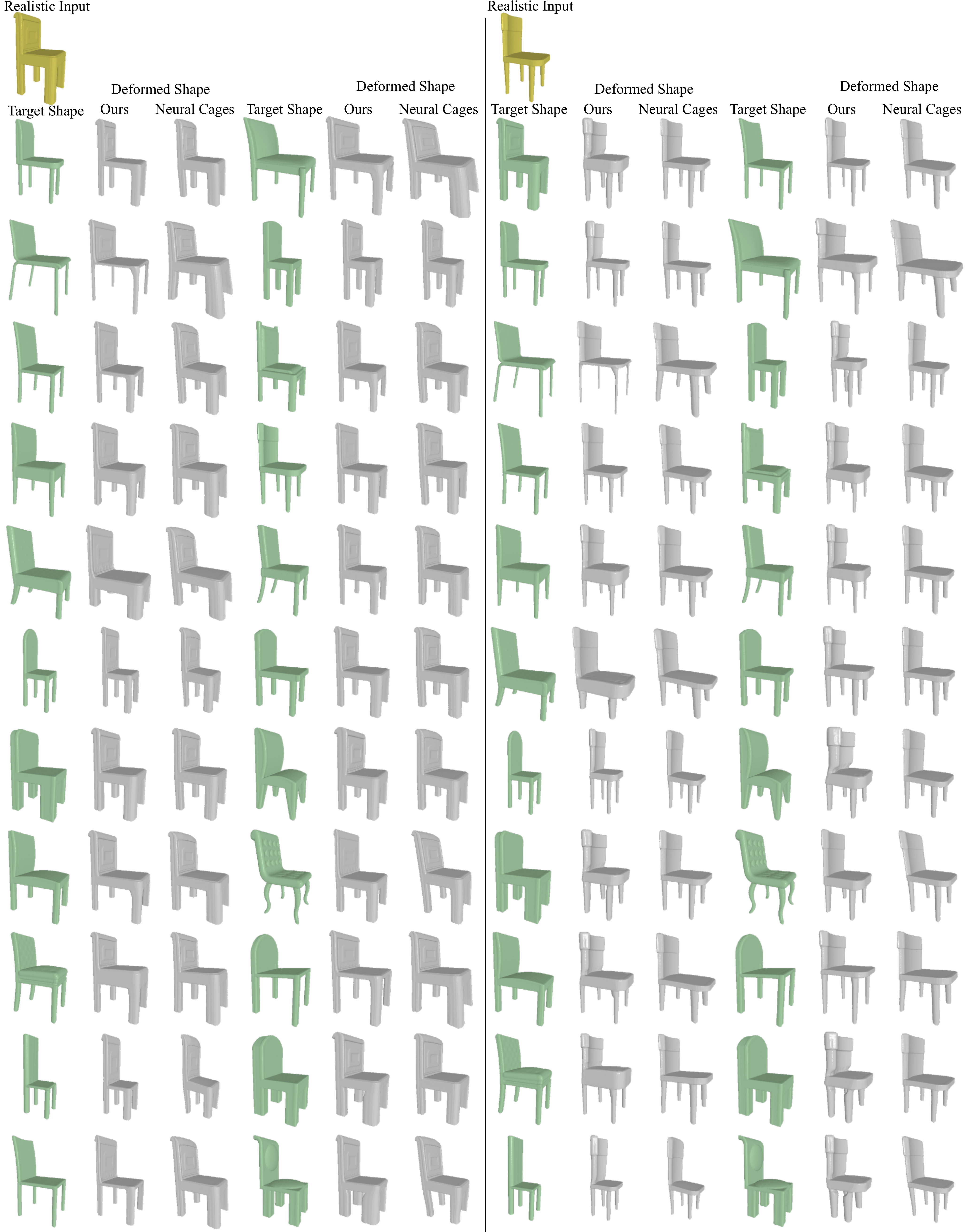}
\caption{\textbf{Additional comparison to Neural Cages~\cite{yifan2019neural} on chairs.} Each source shape (yellow) is deformed to match 22 target shapes (green), using both the proposed method and Neural Cages. Both methods are able to globally match the target shape, but Neural Cages often exhibits distortion in regions of large deformation and cannot match semantic parameters if local deformation is required. In contrast, our method supports more accurate and granular manipulation.}
\label{fig:comp-neuralcage-chair2}
\end{figure*}

\begin{figure*}[t]
   \centering
   \includegraphics[width=0.86\linewidth]{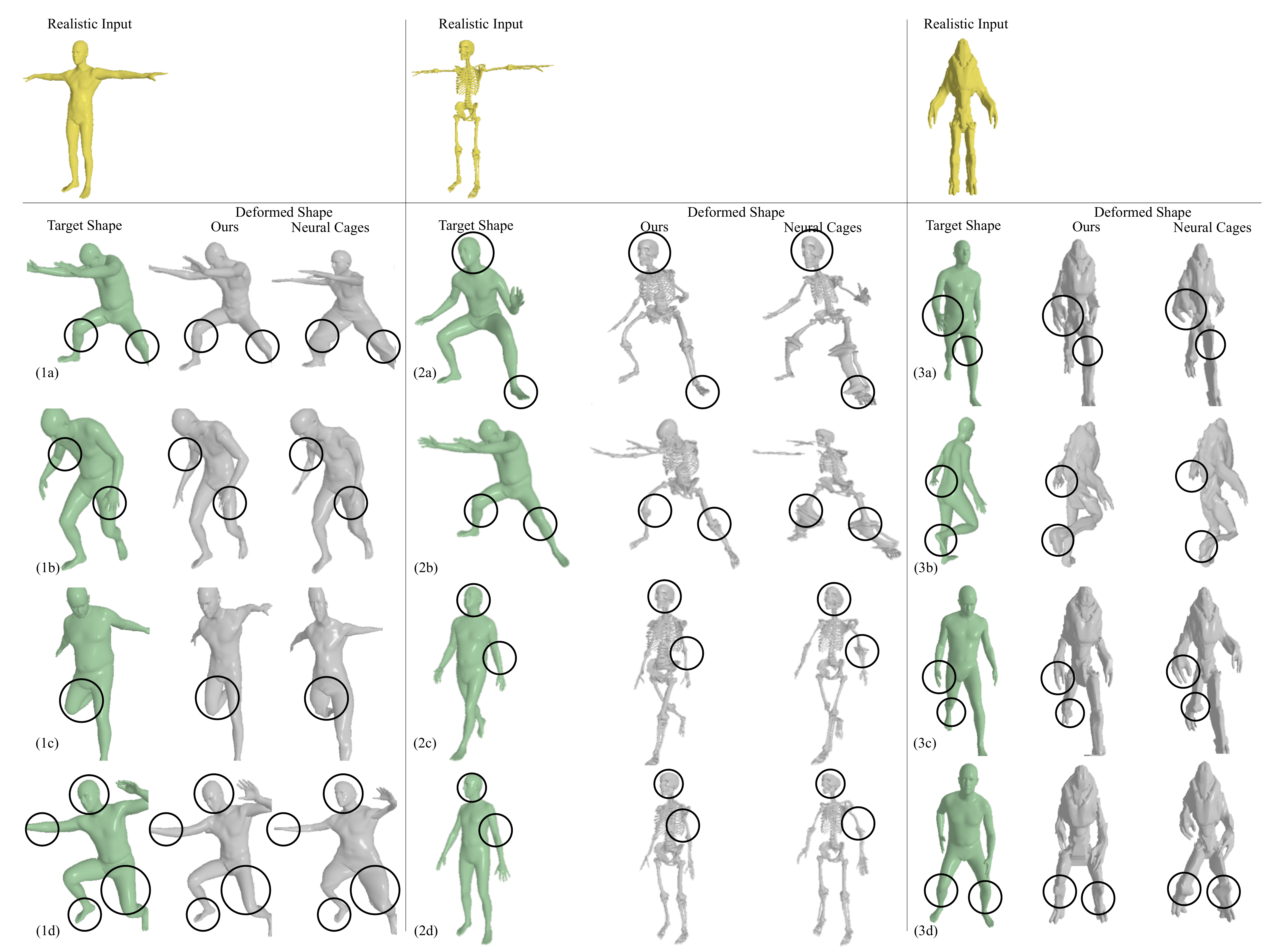}
\caption{\textbf{Additional Comparison to Neural Cages~\cite{yifan2019neural} on human bodies.} Each source shape (yellow) is deformed to match four target shapes (green), using both the proposed method and Neural Cages. Both methods are able to globally match the target shape, but Neural Cages often exhibits distortion in regions of large deformation and cannot match semantic parameters if local deformation is required. In contrast, our method support more accurate and granular manipulation, even for extreme poses.}
\label{fig:comp-neuralcage-human}
\end{figure*}

\begin{figure*}[t!]
   \vspace{-0.5cm}
   \centering
   \includegraphics[width=0.75\linewidth,trim={0cm 0cm 0cm 0cm},clip]{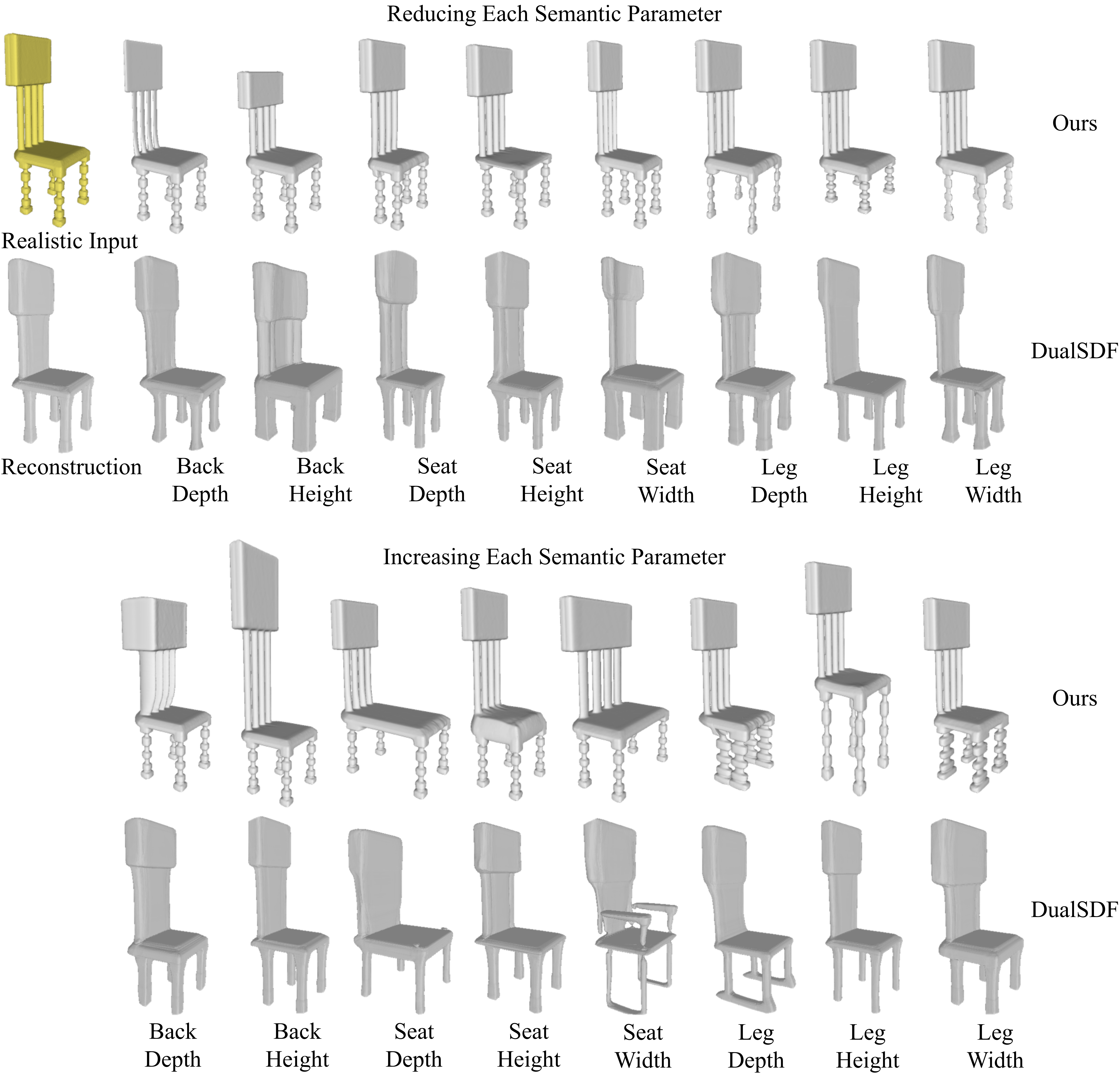}\\
   \vspace{0.6cm}
   \includegraphics[width=0.92\linewidth,trim={0cm 0cm 0cm 0cm},clip]{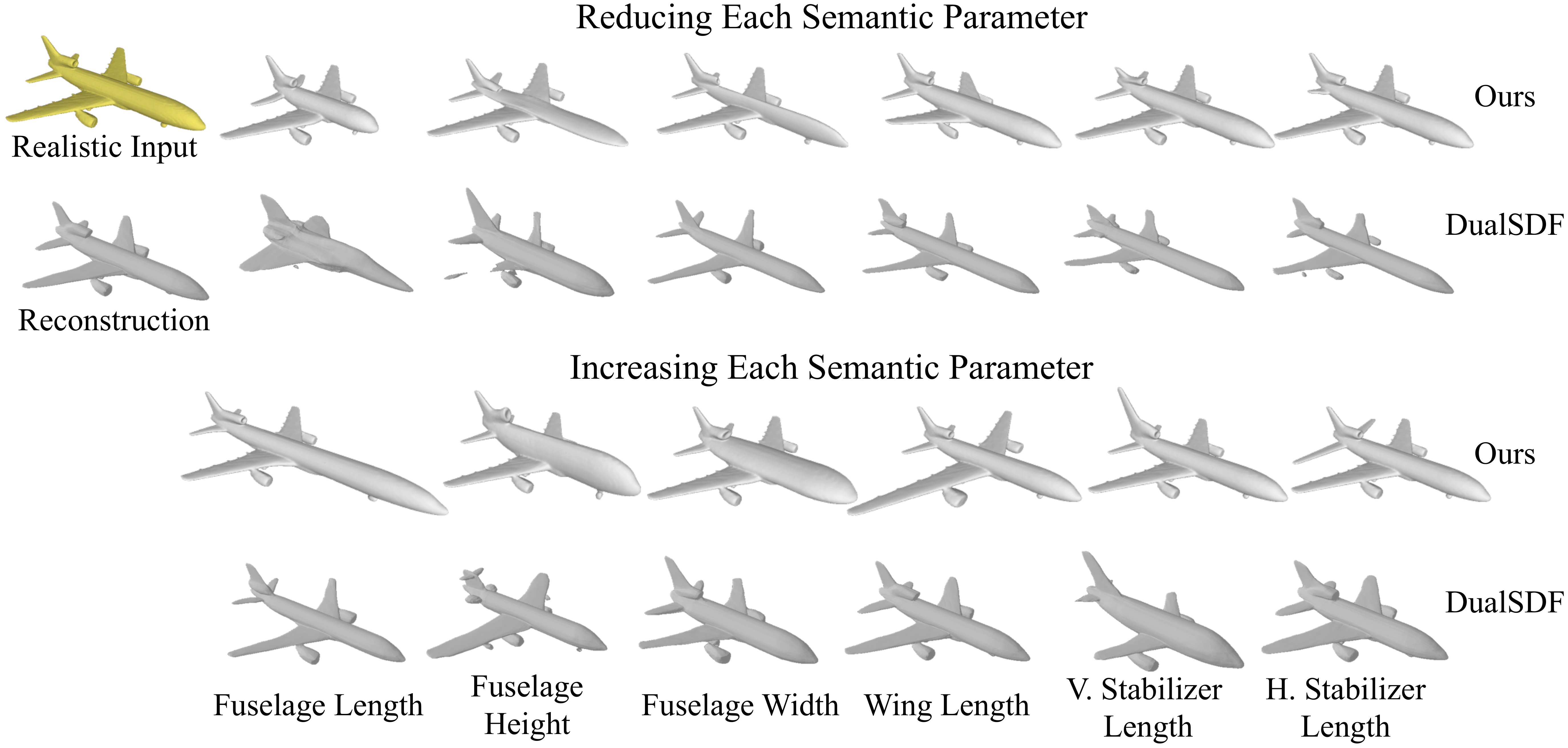}
   \caption{\textbf{Comparison to DualSDF~\cite{hao2020dualsdf}.} The input (yellow) is edited by changing a semantic parameter in our system, or adjusting the radius of a primitive in DualSDF.  When changing a local parameter with DualSDF, the global shape is also affected, as seen in the arms that incorrectly show up in the result of ``increasing seat width''. Also, in contrast to the proposed method, the DualSDF results do not preserve details such as the poles on the chair back and the landing gear and flap track fairings on the back of the wings for the airplane.  }
\label{fig:sup-comp-dualsdf}
\end{figure*}

To compare against Neural Cages~\cite{yifan2019neural}, which was designed for the task of source-to-target deformation rather than direct manipulation of semantic parameters, we modify the proposed method to use a deformation determined by synthetic estimates from both the source and target shapes. Fig.~\ref{fig:comp-neuralcage-chair1} and Fig.~\ref{fig:comp-neuralcage-chair2} show examples using four different source chairs and 22 target chairs. Fig.~\ref{fig:comp-neuralcage-human} shows all three human body shapes evaluated in and with templates provided by the original Neural Cages work. All three human body examples are not seen during the training of the proposed method.
Both methods are able to match the target shape globally. However, note the spurious deformation in the legs in the Neural Cages results as observed in Fig.~\ref{fig:comp-neuralcage}, for example, the 6th, 10th, and last deformed shapes for the first source chair, as well as the fact that in most examples in Fig.~\ref{fig:comp-neuralcage-chair1} and Fig.~\ref{fig:comp-neuralcage-chair2}, the Neural Cages result matches the seat thickness of the source, not the target. In contrast, the proposed method does not introduce unnecessary deformations, yet is able to deform the source locally, not just globally, to match the seat thickness. In the human body results in Fig.~\ref{fig:comp-neuralcage-human}, Neural Cages can introduce distortions in regions with large deformation, such as the knees in example (1a), (1c), (2b) and (3d) or disproportionately scaled hands and feet in examples (1d) (2a) (3a) (3b) and (3c). We observe that, compared to deformation methods based on coarse geometric handles, our results support more granular manipulation even in extreme poses.


We also compare the proposed method with DualSDF~\cite{hao2020dualsdf} on the task of part manipulation. We consider examples common to DualSDF's training set and our testing set. To generate results for DualSDF, we first obtain the embedding of an input shape and then manually adjust the radius of a primitive on the corresponding part along the corresponding dimension. Fig.~\ref{fig:sup-comp-dualsdf} shows the results for chairs and airplanes, the two classes shared by DualSDF and our method. The DualSDF reconstruction, which requires re-synthesis, loses details of the original shape \eg, the poles on the back and the joints on the legs for the chair, the landing gear and flap track fairings on the back of the wings for the airplane. In addition, manipulating a single primitive in one part will cause deformation in other parts as well, since the latent embeddings from DualSDF are highly entangled. For example, when manipulating primitives on the fuselage (the first two airplane examples), the wings are also changed by DualSDF, but correctly preserved by the proposed method. 

\end{document}